\documentclass[sigconf]{acmart}

\usepackage{booktabs} 
\usepackage{t1enc}

\newif\ifDoubleBlind
\DoubleBlindfalse

\usepackage{pgfplots}
\usepackage{placeins}
\usepackage{tikz}
\usepackage{amsfonts}

\newcommand{\abs}[1]{\left| #1\right|}

\newcommand{\set}[1]{\left\{ #1\right\}}
\newcommand{\gilt}{:}
\newcommand{\sodass}{\,:\,}
\newcommand{\setGilt}[2]{\left\{ #1\sodass #2\right\}}

\newcommand{\realrange}[2]{\left[#1, #2\right]}

\newcommand{\unitrange}[2]{\realrange{0}{1}}

\newcommand{\llabel}[1]{\label{\labelprefix:#1}}
\newcommand{\labelprefix}{} 

\newcommand{\discussionsize}{\small}

\newcommand{\frage}[1]{}

\newenvironment{code}{\noindent
\begin{tabbing}%
\hspace{2em}\=\hspace{2em}\=\hspace{2em}\=\hspace{2em}\=\hspace{2em}\=%
\hspace{2em}\=\hspace{2em}\=\hspace{2em}\=\hspace{2em}\=\hspace{2em}\=%
\kill}{\end{tabbing}}

\newcommand{\labelcommand}{}
\newcommand{\captiontext}{}
\newsavebox{\codeparam}
\newcounter{lineNumber}
\newenvironment{disscodepos}[3]{%
\renewcommand{\labelcommand}{#2}%
\renewcommand{\captiontext}{#3}%
\sbox{\codeparam}{\parbox{\textwidth}{#3}}%
\begin{figure}[#1]\begin{center}\begin{code}\setcounter{lineNumber}{1}}{%
\end{code}\end{center}\caption{\llabel{\labelcommand}\captiontext}\end{figure}}

{\end{disscodepos}}

\newcommand{\Is}       {:=}

\newdimen\endofsize\endofsize=0.5em
\def\endofbeweis{~\quad\hglue\hsize minus\hsize
                 \hbox{\vrule height \endofsize width
\endofsize}\par}

\usepackage{epsfig}
\usepackage{url}
\usepackage{amsmath}
\usepackage{amssymb}
\usepackage{pdflscape}
\usepackage{latexsym}
\usepackage{todonotes}
\usepackage{hyperref}
\usepackage{float}
\usepackage{algorithmic}
\usepackage{algorithm}
\usepackage{multicol}
\usepackage{wrapfig}
\usepackage{booktabs}
\usepackage{numprint}
\npdecimalsign{.} 
\usepackage{theorem}

\newcommand{\ie}{i.e.\ }

\usepackage{color}
\usepackage{etoolbox}
\usepackage{geometry}
\newcommand{\csch}[1]{\color{orange}{\small [CS: #1]}\color{black}}
\newcommand{\djs}[1]{\color{blue}{\small [DS: #1]}\color{black}}
\renewcommand{\csch}[1]{}
\renewcommand{\djs}[1]{}

\newcommand{\mytitle}{Distributed Evolutionary $k$-way Node Separators}

\acmConference[]{}{}{}

\begin{document}

\title{\mytitle}

\ifDoubleBlind
\else 
\titlenote{This work was partially supported by DFG grants SA 933/11-1.}
\fi{}

\author{Peter Sanders}
\affiliation{%
  \institution{Karlsruhe Institute of Technology}
  \city{Karlsruhe} 
  \state{Germany} 
}
\email{sanders@kit.edu}

\author{Christian Schulz}
\affiliation{%
  \institution{Karlsruhe Institute of Technology}     \city{Karlsruhe} 
  \state{Germany} 
and \institution{University of Vienna}
  \city{Vienna} \state{Austria} 
}
\email{christian.schulz@univie.ac.at}

\author{Darren Strash}
\affiliation{%
  \institution{Colgate University}     \city{Hamilton} 
  \state{USA} 
}
\email{dstrash@colgate.edu}
\author{Robert Williger}
\affiliation{%
  \institution{Karlsruhe Institute of Technology}
  \city{Karlsruhe} 
  \state{Germany} 
}
\email{williger@ira.uka.de}
\begin{abstract}
Computing high quality node separators in large graphs is necessary for a variety of applications, ranging from divide-and-conquer algorithms to VLSI design. 
In this work, we present a novel distributed evolutionary algorithm tackling the $k$-way node separator problem.
A key component of our contribution includes new $k$-way local search algorithms based on maximum flows. We combine our local search with a multilevel approach to compute an initial population for our evolutionary algorithm, and further show how to modify the coarsening stage of our multilevel algorithm to create effective combine and mutation operations.
Lastly, we combine these techniques with a scalable communication protocol, producing a system that is able to compute high quality solutions in a short amount of time.
Our experiments against competing algorithms show that our advanced evolutionary algorithm computes the best result on 94\% of the chosen benchmark instances. 
\end{abstract}


\maketitle

\section{Introduction}
Given a graph $G=(V,E)$, the \emph{$k$-way node separator problem} is to find a small \emph{separator} $S\subset V$, and $k$ disjoint subsets of $V$, $V_1, \ldots, V_k$ called \emph{blocks}, such that no edges exist between two different blocks $V_i$ and $V_j$ ($i \neq j$) and $V= \bigcup_i V_i \cup S$. The objective is to minimize the size of the separator $S$ or, depending on the application, the cumulative weight of its nodes, while the blocks $V_i$ are balanced. 
Note that removing the set $S$ from the graph results in at least~$k$~connected~components.

Many algorithms rely on small node separators. 
For example, small balanced separators are a popular tool in divide-and-conquer strategies~\cite{lipton1980applications,leiserson1980area,BHATT1984300}, are useful to speed up the computations of shortest paths~\cite{schulz2002using,delling2009high,dibbelt2014customizable}, are necessary in scientific computing to compute fill reducing orderings with nested dissection algorithms~\cite{george1973nested}~or~in~VLSI~design~\cite{leiserson1980area,BHATT1984300}.

Finding a \emph{balanced} node separator is NP-hard for general graphs even if the maximum node degree is three~\cite{bui1992finding,garey2002computers}. 
Therefore, heuristic and approximation algorithms are used in practice to find small node separators.
The most commonly used method to solve the node separator problem on large graphs in practice is the \emph{multilevel} approach.
During a coarsening phase, a multilevel algorithm reduces the graph size by iteratively contracting the nodes and edges of $G$ until the graph is small enough to compute a node separator by some other (presumably time consuming) algorithm. A node separator of the input graph is then constructed by iteratively uncontracting the graph, transferring the solution to this finer graph, and then applying local search algorithms to improve the solution. 

Although current solvers are typically fast enough for most applications, they unfortunately produce separators of low solution quality. This may be acceptable in applications that use a separator just once, however, many applications first compute a separator as a preprocessing step, and then rely on a high-quality separator for speed in later stages. This is true in VLSI design~\cite{leiserson1980area,BHATT1984300}, where even small improvements in separator size can have a large impact on computation time and production costs. High-quality node separators can also speed up shortest path queries in road networks, for example, in customizable contraction hierarchies~\cite{dibbelt2014customizable}, where smaller node separators yield better node orderings that are repeatedly used to answer shortest path queries. The cost for computing one high quality node separator is then amortized over a many shortest path queries.  
Hence, our focus is on solution quality in this work.

\subsection{Our Results}
The main contribution of this paper is a technique that integrates an evolutionary search algorithm with a novel multilevel $k$-node separator algorithm and its scalable parallelization. 
We present novel mutation and combine operators for the problem which are based on the multilevel scheme. 
Due to the coarse-grained parallelization, our system is able to compute separators that have high quality \textit{within a few minutes} for graphs of moderate size. 
\vfill
\pagebreak

\section{Preliminaries}
\label{s:preliminaries}
\subsection{Basic Concepts}
Throughout this paper, we consider an undirected graph $G=(V=\{0,\ldots, n-1\},E)$ with $n = |V|$, and $m = |E|$.
$\Gamma(v)\Is \setGilt{u}{\set{v,u}\in E}$ denotes the \emph{neighborhood} of a node $v$.
A graph $S=(V', E')$ is said to be a \emph{subgraph} of $G=(V, E)$ if $V' \subseteq V$ and $E' \subseteq E \cap (V' \times V')$. We call $S$ an \emph{induced} subgraph when $E' = E \cap (V' \times V')$.  
For a set of nodes $U\subseteq V$, $G[U]$ denotes the subgraph induced by $U$.

The \emph{graph partitioning problem}, which is closely related to the node separator problem, asks for \emph{blocks} of nodes $V_1$,\ldots,$V_k$ 
that partition $V$ (i.e., $V_1\cup\cdots\cup V_k=V$ and $V_i\cap V_j=\emptyset$
for $i\neq j$). A \emph{balancing constraint} demands that 
$\forall i\in \{1..k\}\gilt |V_i|\leq L_{\max}\Is (1+\epsilon)\lceil |V|/k \rceil$ for
some parameter $\epsilon$. 
In this case, the objective is often to minimize the total \emph{cut} $\sum_{i<j}|E_{ij}|$ where 
$E_{ij}\Is\setGilt{\set{u,v}\in E}{u\in V_i,v\in V_j}$. 
The set of cut edges is also called \emph{edge separator}.
An abstract view of the partitioned graph is the so called \emph{quotient graph}, where nodes represent blocks and edges are induced by~connectivity~between~blocks. 

The \emph{node separator problem} asks to find blocks, $V_1, \ldots, V_k$ and a separator $S$ that partition~$V$ such that there are no edges between the blocks. 
Again, a balancing constraint demands $|V_i| \leq (1+\epsilon)\lceil|V|/k \rceil $. However, there is no balancing constraint on the separator~$S$. 
The objective is to minimize the size of the separator $|S|$. 
Note that removing the set $S$ from the graph results in at least $k$ connected components and that the blocks $V_i$ itself do not need to be connected components.
Two blocks of $V_{i}$ and $V_{j}$ are \textit{adjoint} if there exists a separator node $s \in S$ that connects both blocks. 
Note that $s$ can separate more than two blocks. 
For the separator case, edges in the \emph{quotient graph} are induced by adjoint blocks.
By default, our initial inputs will have unit edge and node weights. However, the results in this paper are easily transferable to node and edge~weighted~problems.

A matching $M\subseteq E$ is a set of edges that do not share any common nodes,
\ie the graph $(V,M)$ has maximum degree one.  \emph{Contracting} an edge $\set{u,v}$ replaces the nodes $u$ and $v$ by a
 new node $x$ connected
to the former neighbors of $u$ and $v$. We set $c(x)=c(u)+c(v)$.
If replacing
edges of the form $\set{u,w},\set{v,w}$ would generate two parallel edges
$\set{x,w}$, we insert a single edge with
$\omega(\set{x,w})=\omega(\set{u,w})+\omega(\set{v,w})$.
\emph{Uncontracting} an edge $e$ ``undoes'' its contraction. 
In order to avoid tedious notation, $G$ will denote the current state of the graph
before and after a (un)contraction unless we explicitly want to refer to 
different~states~of~the~graph.

The multilevel approach consists of three main phases.
In the \emph{contraction} (coarsening) phase, 
we iteratively identify matchings $M\subseteq E$ 
and contract the edges in $M$. Contraction should quickly reduce the size of the input and each computed level
should reflect the global structure of the input graph. 
Contraction is stopped when the graph is small enough so that the problem can be solved by some other potentially more expensive algorithm. 
In the \emph{local search} (or uncoarsening) phase, matchings are iteratively uncontracted.  
After uncontracting a matching, a local search algorithm moves nodes to decrease the size of the separator or to to improve balance of the block while keeping the size of the separator. 
The intuition behind the approach is that a good solution at one level of the hierarchy will also be a good solution on the next finer level so that local search will quickly find a good solution.

\subsection{Related Work}
\label{s:related}
\csch{update this section}
Here, we focus on results closely related to our main contributions, as well as previous work on the node separator problem. However, we briefly mention that there has been a \emph{huge} amount of research on graph partitioning, which is closely related to the node separator problem. We refer the reader to \cite{GPOverviewBook,SPPGPOverviewPaper} for thorough reviews of the results in this area. 

\paragraph{$2$-way Node Separators.}
In contrast to the NP-hardness of the problem in general, Lipton and Tarjan~\cite{lipton1979separator} showed that small $2$-way balanced separators can always be found in linear time for planar graphs. Their \textit{planar separator theorem} states that, for planar graphs, one can always find a $2$-way separator $S$ in linear time such that $|S| = O(\sqrt{|V|})$ and $|V_i| \leq 2 |V| / 3$. Note that, to achieve better balance, the problem remains NP-hard~\cite{fukuyama2006np} even for planar graphs. \djs{state most concretely}

For general graphs there are several heuristics to compute small node separators.
A common and simple method is to first compute an edge separator using a multilevel graph partitioning algorithm, and then compute a node separator by selecting nodes incident to the edge separator~\cite{pothen1990partitioning,dissSchulz}.
A major drawback to this method is that the graph partitioning objective---to minimize the number of cut edges---differs from the objective of the node separator problem. This difference in combinatorial structure, unfortunately, means that graph partitioning approaches are unlikely to find~high~quality~solutions. 

The Metis~\cite{karypis1998fast} and Scotch~\cite{scotch} graph partitioners use a multilevel approach to obtain a $2$-way node separator.  After contraction, both tools compute a node separator on the coarsest graph using a greedy algorithm.
This separator is then transferred level-by-level, dropping non-needed nodes on each level and applying Fiduccia-Mattheyses (FM) style local search.  LaSalle and Karypis~\cite{lasalle2015efficient} further gave a shared-memory parallel algorithm 
and showed that a multilevel approach combining greedy local search with a segmented FM algorithm can outperform serial FM algorithms.


Other recent approaches look at variations of the node separator problem, such as Pareto solutions for edge cut versus balance~\cite{hamann2015graph}, and enforcing both upper \emph{and} lower bounds on block sizes~\cite{hager2014multilevel}.

\paragraph{$k$-way Node Separators.} In the theoretical algorithms literature, a \emph{multiway} separator is the generalization of separator to a higher number of blocks.  Frederickson~\cite{frederickson-1987} first gave this generalization, and showed that planar graphs can be partitioned into $O(|V|/k)$ subsets of size at most $k$, where each subgraph shares $O(\sqrt{k})$ boundary vertices with other subgraphs. Frederickson's generalization of a separator is different than what we consider here: we do not attempt to minimize the number of nodes that separate pairs of blocks, instead we minimize the total number of nodes that collectively separate all blocks. In practice, the balance constraint on the block size may also implicitly minimize the number of separator nodes between blocks, but this is not strictly enforced.

We are currently unaware of any existing algorithm to compute a $k$-way node separator. However, we note that, similar to $2$-way node separators, a $k$-way node separator may be computed by first computing a $k$-way \emph{edge} separator, and then keeping some or all of the nodes incident to the edge separator as a node separator.

\subsection{Detailed Related Work}
\label{s:detailedrw}
\ifDoubleBlind
\else
Recently, we presented a new multilevel algorithm for the 2-way node separator problem~\cite{2waynodeseparators}. We outline the details of the multilevel algorithm here since we modify this algorithm to compute $k$-way node separators and then use the modified algorithm to compute the initial population of the evolutionary algorithm.
\fi{} 

During coarsening, they use a two-phase approach, which makes contraction more systematic by separating two issues: A \emph{rating function} and a \emph{matching} algorithm. 
Hence, the coarsening algorithm captures both \emph{local} information and the \emph{global} structure of the graph. 
While the rating function allows a flexible characterization of what a ``good'' contracted graph is, the simple, standard definition of the matching problem allows to reuse previously developed algorithms for weighted matching. 
Their method further uses the \textit{Global Path Algorithm (GPA)}~\cite{MauSan07} to compute matchings, which runs in near-linear time. 
GPA scans the edges in order of decreasing weight (rating)
but rather than immediately building a matching, it first constructs a collection
of paths and even-length cycles. Afterwards, \textit{optimal solutions} are computed for each
of these paths and cycles using~dynamic~programming. 

Coarsening is stopped as soon as the graph has less than ten thousand nodes. 
To compute an initial separator, Sanders and Schulz compute an edge separator, from which they derive a node separator. 
Conversion is done by using all boundary nodes as initial separator $S$ and using the \emph{flow-based technique} described below to select the smallest separator contained in the induced~bipartite~subgraph. 

Two-way local search is based on localized local search and flow-based techniques.
Localized local search algorithms for the node separator problem are initialized only with a subset of given separator instead of the whole separator set. 
In each iteration a separator node with the highest priority not violating the balance constraint is moved.
The priority is based on the \emph{gain} concept, \ie the decrease in the objective when the separator node is moved into a block. 
More precisely, the gain of the node is the weight of $v$ minus the weight of the nodes that have to be added to the separator once it is moved. 
Each node is moved at most once out of the separator within a single local search. 
After a node is moved, newly added separator nodes become eligible for movement.

The other local search algorithm contributed is based on maximum flows.
More precisely, the authors solve a node-capacitated flow problem $\mathcal{F}=(V_\mathcal{F}, E_\mathcal{F})$ to improve a given node separator.
We shortly outline the details.
Given a set of nodes $A \subset V$, its \emph{border} is defined as
$\partial A := \{ u \in A \mid \exists (u,v) \in E : v \not\in A\}$.
The set $\partial_1 A := \partial A \cap V_1$ is called \emph{left border} of $A$ and the set $\partial_2 A := \partial A \cap V_2$ is called \emph{right border} of $A$. 
An \emph{$A$ induced flow problem} $\mathcal{F}$ is the node induced subgraph $G[A]$ using $\infty$ as edge-capacities and the node weights of the graph as node-capacities. Additionally there are two nodes $s,t$ that are connected to the border of $A$. 
More precisely, $s$ is connected to all left border nodes $\partial_1 A$ and all right border nodes $\partial_2 A$ are connected to $t$.  
These new edges get capacity $\infty$. 
Note that the additional edges are directed.
$\mathcal{F}$ has the \emph{balance property} if each ($s$,$t$)-flow induces a balanced node separator in $G$, \ie the blocks $V_i$ fulfill the balancing constraint.
The basic idea is to construct a flow problem $\mathcal{F}$ having the balance property. 
Such a subgraph is found by performing breadth first searches (BFS) initialized with the separator nodes.

The algorithms starts by setting $A$ to $S$ and extending it by performing two BFS. 
The first BFS is initialized with the current separator nodes $S$ and only looks at nodes in block $V_1$.
The same is done during the second BFS with the difference that now looks at nodes of block $V_2$.
Each node touched by any of the BFS is added to $A$.
The BFSs are stopped in such a way that the final solution of the flow problem (which can be transformed into a separator of the original problem) yields a balanced separator in the original graph.
To obtain even better solutions, larger flow-problems can be defined by dropping the requirement that each cut in the flow problem corresponds to a \emph{balanced} node separator in the original graph. If the resulting node separator is not balanced, then the algorithms starts again with a smaller flow problems, \ie stopping the BFS to define the flow problem earlier.

To improve solution quality, the notion of \emph{iterated multilevel schemes} has been introduced to the node separator problem.
Here, one transfers a solution of a previous multilevel cycle down the hierarchy and uses it as initial solution. 
More precisely, this can be done by not contracting any cut edge, \ie an edge running between a block and the separator. 
This is achieved by modifying the matching algorithm to not match any edge that runs between $V_i$ and $S$ ($i=1,2, \ldots, k$). 
Hence, when contraction is done, every edge leaving the separator will remain and one can transfer the node separator down in the hierarchy.
Thus a given node separator can be used as initial node separator of the coarsest graph (having the same balance and size as the node separator of the finest graph).  
This ensures non-decreasing quality, if the local search algorithm guarantees no worsening. 
\begin{figure}[b]
\centering
    \includegraphics[page=1,width=.23\textwidth]{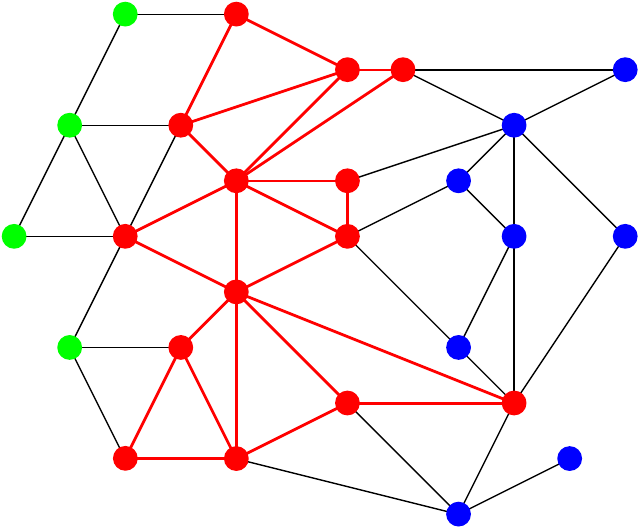} 
    \includegraphics[page=2,width=.23\textwidth]{plots/Preprocessing}
    \caption{Illustration of the preprocessing for a graph with blocks $V_{1}$ (\textcolor{green}{green}), $V_{2}$ (\textcolor{blue}{blue}) and separator $S$ (\textcolor{red}{red}).}
    \label{fig:preprocessing}
\end{figure}

\section{Local Search and Balancing}
\label{s:localsearch}
Local search is used to improve a given solution to the node separator problem on each level of the multilevel hierarchy. As the term implies we locally search for a better separator around the current separator in order to find a new locally optimal solution. In this section, we show how we can use the flow-based local improvement method described above to locally improve a $k$-way separator for $k>2$ and describe algorithms that are able to balance solutions, e.g. solutions that contain blocks with too many vertices. These algorithms are used to create the initial population of our evolutionary algorithm as well as to provide the combine and mutation operations.
Our algorithm is called \emph{Adv} and the evolutionary algorithm that is introduced later \emph{AdvEvo}.

\subsection{$k$-way Local Search}
Our $k$-way local search builds on top of the flow-based search which is intended for improving a separator with $k=2$. 
The main idea is to find pairs of \emph{adjoint blocks} and then perform local search on the subgraph induced by adjoint block pairs.

\paragraph{Preprocessing.} In order to find pairs of adjoint blocks, we look at separator nodes which \textit{directly separate} two different blocks, meaning these separator nodes are adjacent to nodes from at least two different blocks not including the separator.
In general directly separating nodes 
do not have to exist (see Figure~\ref{fig:preprocessing}).
In other words, it may be that a separator disconnects two blocks, but the shortest path distance between the blocks is greater or equal to two. 
Using a preprocessing step, we first make sure that each separator node is adjacent to at least two blocks, \ie each separator node~is~directly~separating.

The preprocessing step works as follows: 
we iterate over all separator nodes and try to remove them from the separator if they do not directly separate two blocks.
The order in which we look at the separator nodes is given by the number of adjacent blocks (highest first).
Let $s$ be the current separator node under consideration.
If it has two or more non-separator neighbors in different blocks, it already directly separates at least two blocks and we continue. 
If $s$ only has neighbors in a single block in addition to the separator, we move it into that block. 
Lastly, if $s$ only has other separator nodes as neighbors, we put it into a block having smallest overall weight. 
In each step, we update the priorities of adjacent neighboring~separator~nodes. Note that nodes are only removed from the separator and never added. Moreover, removing a node from the separator can increase the priority of an adjacent separator node only by one. As soon as the priority of a node is larger than one, it is directly separating and we do not have to look at the vertex again. After the algorithm is done, each separator node is directly separating at least two blocks and we can build the quotient graph in order to find adjoint blocks. Our preprocessing can introduce imbalance to the solution. Hence, we run the balance routine defined below after preprocessing. 

\paragraph{Pair-wise Local Search.}\label{pair refinement}
Subsequent to the preprocessing step, we identify the set of all adjoint block pairs $P$ by iterating through all separator nodes and their adjacent nodes. We iterate through all pairs $p = (A, B) \in P$ and build the subgraph $\mathcal{G}_p$. $\mathcal{G}_{p}$ is induced by the set of nodes consisting of all nodes in $A$ and $B$ as well as all separator nodes that \textit{directly separate} the blocks. After building $\mathcal{G}_{p}$, we run local search designed for $2$-way separators on this subgraph. Note that an improvement in the separator size between $A$ and $B$ directly corresponds to an improvement to the overall~separator~size.

To gain even smaller separators and because the solution is potentially modified by local search, we repeat local search multiple times in the following way. 
The algorithm is organized in rounds. In each round, we iterate over the elements in $P$ and perform local search on each induced subgraph.  
If local search has not been successful, we remove $p$~from~$P$. Otherwise, we keep $p$ for~the~next~round. 
\subsection{Balancing}
\begin{figure}[t]
	\centering
	\includegraphics[page=1,width=4.5cm]{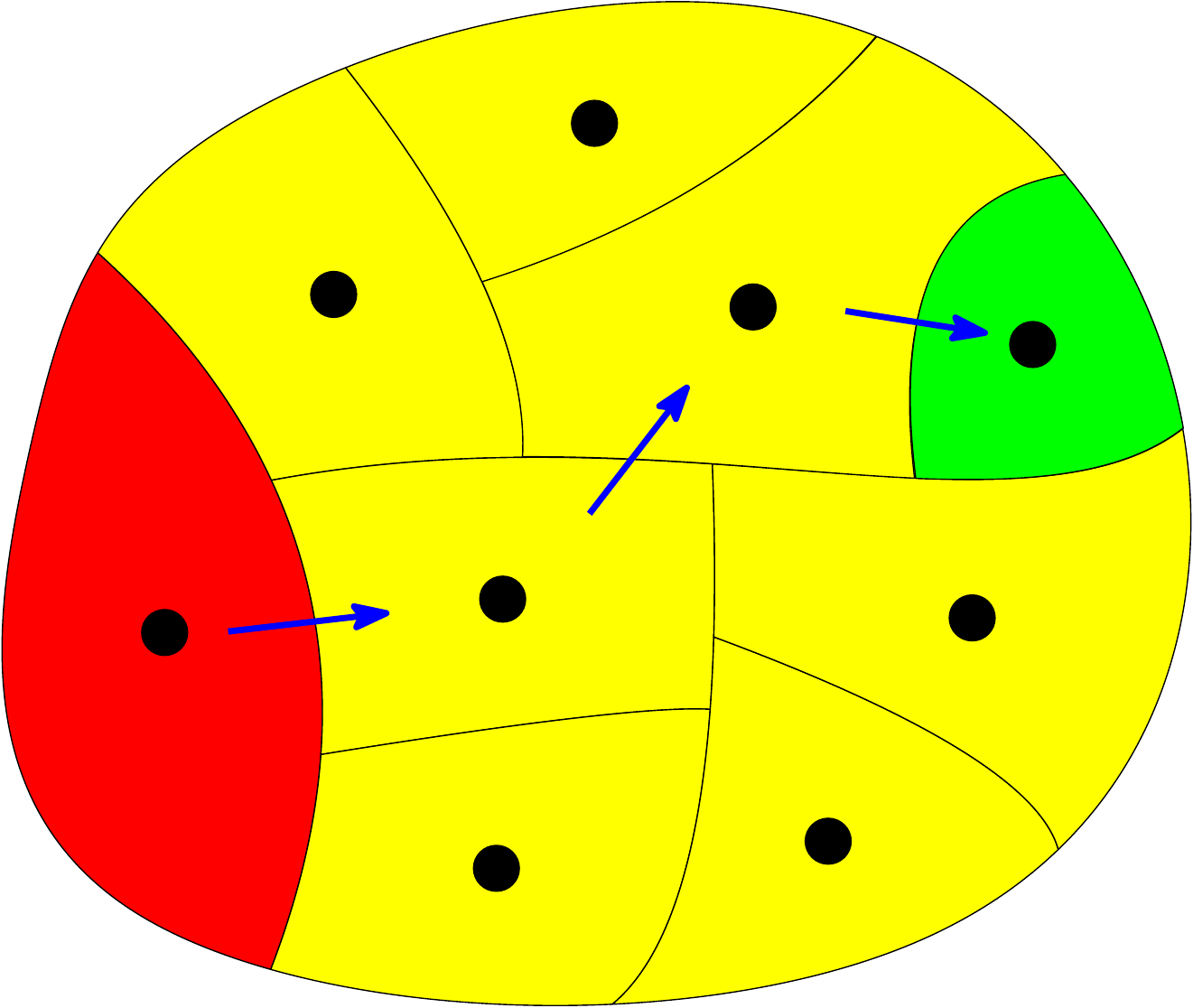}
	\caption{Illustration of the quotient graph and the balancing path (\textcolor{blue}{blue}) from an imbalanced block (\textcolor{red}{red}) to the lightest block (\textcolor{green}{green}).}\label{fig:balancing}
\end{figure}

To guarantee the balance constraint, we use a balance operation. 
Given an imbalanced separator of a graph, the algorithm returns a balanced node separator. 
Roughly speaking, we 
move nodes from imbalanced blocks towards the blocks that can take nodes without becoming overloaded (see Figure~\ref{fig:balancing} for an illustrating example). 
As long as there are imbalanced blocks, we iteratively repeat the following steps:

First, we find a path $p$ from the heaviest block $\mathcal{H}$ to the lightest block $\mathcal{L}$ in the quotient graph. 
If there is no such path, we directly move a node to the lightest block and make its neighbors separator nodes. 
Next, we iterate through the edges $(A, B) \in p$ and move nodes from $A$ into $B$.
In general, we move $\min({L_\text{max}-|\mathcal{L}|, |\mathcal{H}|-L_\text{max}})$ nodes along the path, \ie as many nodes as the lightest block can take without getting overloaded and as little nodes necessary so that the heaviest block is balanced. 
Moving nodes is based on gain of the movement as defined in Section~\ref{s:detailedrw}. 
Basically, we use a priority queue of separator nodes that directly separate $A$ and $B$ with the key being set to the gain.
Note that these movements create new separator nodes and can potentially worsen solution quality.
We use the gain definition because our primary target is to minimize the increase~of~the~separator~size.  

Then we dequeue nodes from the our priority queue until $A$ is balanced. 
We move each dequeued node $s$ to $B$ and move its neighbors being in $A$ into the separator and the priority queue. Also the priorities of the nodes in the queue are updated.
After moving the nodes $A$ will be balanced.
If $B$ is imbalanced, we continue with the next pair in the path, \ie sending the same amount of nodes. 
If $B$ is also balanced, we are done with this path and do not move any more nodes.
Our algorithm continues with the next~imbalanced~block. 

\section{Evolutionary $k$-way Node Separators}
\label{s:evolutionarycomponents}
Our EA starts with a population of individuals (in our case the node separator of the graph) and evolves the population into different populations over several rounds. 
In each round, the EA uses a selection rule based on the fitness of the individuals (in our case the size of the separator) of the population to select good individuals and combine them to obtain improved offspring. 
Note that we can use the size/weight of the separator as a fitness function since our algorithm always generates separators fulfilling the given balance constraint, i.e. there is no need to use a penalty function to ensure that the final separator is feasible. 
When an offspring is generated an eviction rule is used to select a member of the population and replace it with the new offspring. 
In general one has to take both into consideration, the fitness of an individual and the distance between individuals in the population~\cite{baeckEvoAlgPHD96}. 
Our algorithm generates only one offspring per generation. 

\subsection{Combine Operation} \label{s:combineoperators}
We now describe the combine operator. 
Our combine operator assures that the offspring has an objective \textit{at least as good as the best of both parents}.  
Roughly speaking, the combine operator combines an individual/separator $\mathcal{P} = V^\mathcal{P}_1, ..., V^\mathcal{P}_k, S^\mathcal{P}$ (which has to fulfill a balance constraint) with a second individual/separator $\mathcal{C} = V^\mathcal{C}_1, ..., V^\mathcal{C}_{k}, S^{\mathcal{C}} $. Let $\mathcal{P}$ be the individual with better~fitness.

The algorithm begins with selecting two individuals from the population.
The selection process is based on the tournament selection rule \cite{Miller95geneticalgorithms}, i.e. $P$ is the fittest out of two random individuals $R_1, R_2$ from the population and the same is done to select $\mathcal{C}$. 
Both node separators are used as input for our multi-level algorithm in the following sense. 
Let $\mathcal{E}$ be the set of edges that are cut edges, i.e. edges that run between blocks and the separator, in either $\mathcal{P}$ \textit{or} $\mathcal{C}$. 
All edges in $\mathcal{E}$ are blocked during the coarsening phase, i.e. they \textit{are not contracted} during the coarsening phase.
In other words these edges are not eligible for the matching algorithm used during the coarsening phase and therefore are not part of any matching.

The stopping criterion of the multi-level algorithm is \textit{modified} such that it stops when no contractable edge is left. 
As soon as the coarsening phase is stopped, we apply the separator $\mathcal{P}$ to the coarsest graph and use this as 
initial separator. 
This is possible since we did not contract any edge running between the blocks and the separator in $\mathcal{P}$. 
Note that due to the specialized coarsening phase and this specialized initial phase we obtain a high quality initial solution on a very coarse graph which is usually not discovered by conventional algorithms that compute an initial solution. 
Since our local search algorithms guarantee no worsening of the input solution and we use random tie breaking we can assure non-decreasing quality.
Note that the local search algorithms can effectively exchange good parts of the solution on the coarse levels by moving only a few vertices. 

Also note that this combine operator can be extended to be a multi-point combine operator, i.e. the operator would use $\ell$ instead of two parents. 
However, during the course of the algorithm a sequence of two point combine steps is executed which somehow "emulates" a multi-point combine step. 
Therefore, we restrict ourselves to the case $\ell=2$.
When the offspring is generated we have to decide which solution should be evicted from the current population.
We evict the solution that is \textit{most similar} to the offspring among those individuals in the population that have an objective worse or equal than the offspring itself. 
Here, we define the similarity $\sigma$ of two node separators $S_{1}$ and $S_{2}$ as the cardinality of the symmetric difference of both separators: $\sigma = \abs{(S_{1} \setminus S_{2}) \cup (S_{2} \setminus S_{1})}$. Therefore $\sigma$ denotes the number of nodes contained in one separator but not in the other. 
This ensures some diversity in the population and hence makes the evolutionary algorithm more effective.

\subsection{Mutation Operation}
The mutation operation works similar to the combine operation. The main difference is that there is only one input individual to the multi-level algorithm and that the offspring can be less fit compared to the input individual.
Hence, only edges that run between the blocks and the separator of that individual are not eligible for the matching algorithm. 
This way the input individual can be transferred downwards in the hierarchy.
Additionally, the solution is not used as initial separator but the initial algorithm is performed to find an initial separator. Note however due to the way the coarsening process is defined the input separator is still contained in the coarsest graph.

\section{Putting Things Together and Parallelization}
\label{s:parallelization}
We now explain parallelization and describe how everything is put together to be our full evolutionary algorithm \emph{AdvEvo}. We use a parallelization scheme that has been successfully used in graph partitioning~\cite{kaffpaE}. Each processing element (PE)  basically performs the same operations using different random seeds (see Algorithm~\ref{alg:localview}). 
First we estimate the population size~$\mathcal{S}$: each PE creates an individuum and measures the time $\overline{t}$ spend. 
We then choose $\mathcal{S}$ such that the time for creating $\mathcal{S}$ node separators is approximately $t_{\text{total}}/f$ where the fraction $f$ is a tuning parameter and $t_{\text{total}}$ is the total running time that the algorithm is given to produce a node separator of the graph. 
Each PE then builds its own population, i.e. our multi-level algorithm is called several times to create $\mathcal{S}$ individuals/separators. 
Afterwards the algorithm proceeds in rounds as long as time is left. 
With corresponding probabilities, mutation or combine operations are performed and the new offspring is inserted~into~the~population.

We choose a parallelization/communication protocol that is quite similar to \textit{randomized rumor spreading} \cite{conf/icalp/DoerrF11} which has shown to be scalable in an evolutionary algorithm for graph partitioning~\cite{kaffpaE}. We follow their description closely.
Let $p$ denote the number of PEs used. A communication step is organized in rounds. 
In each round, a PE chooses a communication partner and sends her the currently best node separator $P$ of the local population. 
The selection of the communication partner is done uniformly at random among those PEs to which $P$ not already has been send to.  
Afterwards, a PE checks if there are incoming individuals and if so inserts them into the local population using the eviction strategy described above.
If $P$ is improved, all PEs are again eligible.
This is repeated $\log p$ times.
The algorithm is implemented \textit{completely asynchronously}, i.e. there is no need for a global synchronization.

\begin{algorithm}[h]
\caption{locallyEvolve}
\small
\begin{algorithmic}
\STATE   \quad estimate population size $S$
\STATE   \quad \textbf{while} time left 
\STATE   \quad \quad \textbf{if} elapsed time $< t_{\text{total}}/f$ \textbf{then} 
\STATE   \quad \quad \quad create individual, insert into local population 
\STATE   \quad \quad \textbf{else}
\STATE   \quad \quad\quad flip coin $c$ with corresponding probabilities
\STATE   \quad \quad\quad \textbf{if} $c$ shows head \textbf{then} perform mutation   
\STATE   \quad \quad\quad \textbf{else} perform combine 
\STATE   \quad \quad\quad insert offspring into population if possible 
\STATE   \quad \quad communicate according to comm. protocol
\end{algorithmic}
\label{alg:localview}
\end{algorithm}
\vspace*{-.5cm}
\subsection{Miscellaneous}
Besides \emph{Adv} and \emph{AdvEvo}, we also use two more algorithms to compare solution quality. 
The first one is a sequential algorithm that starts by computing a $k$-way partition using KaFFPa-Strong and derives a $k$-way separator by pair-wise decoupling by using the method of Pothen and Fan~\cite{pothen1990partitioning} on each adjacent pair of blocks. The main idea of Pothen and Fan is to compute a minimum vertex cover in the bipartite subgraph induced by the set of cut edges between two pairs of blocks. The union of the computed separators nodes is a $k$-way separator. In our experiments, the algorithm is called \emph{Simple}.
The second algorithm, is a modification of KaFFPaE~\cite{kaffpaE} which is an evolutionary algorithm to compute graph partitions.
We modify the fitness function to be the size of the separator that can be derived using the \emph{Simple} approach, but keep the rest of the algorithm. 
More precisely, this means that the population of the algorithm are still graph partitions instead of separators, but for example selection is based on the size of the derivable separator.
Additionally, the combine operations in KaFFPaE still optimize for cuts instead of separators.
This algorithm is called \emph{SimpleEvo}.
\section{Experiments}
\begin{table}[b]
	\centering
        \small
	\begin{tabular}{| l | r || l | r | }
			\hline
		 	Graph & $n$ & Graph & $n$ \\
		 	\hline \hline
		 	\multicolumn{2}{|c||}{Walshaw Graphs} &  \multicolumn{2}{c|}{Walshaw Graphs}\\
			\hline

		 	add20       & \numprint{2395} &bcsstk32    & \numprint{44609} \\
		 	data        & \numprint{2851} &fe\_body*   & \numprint{45087} \\
		 	3elt        & \numprint{4720} &t60k        & \numprint{60005} \\
		 	uk          & \numprint{4824} &wing        & \numprint{62032} \\
		 	add32       & \numprint{4960} &brack2      & \numprint{62631} \\
		 	bcsstk33*   & \numprint{8738} &finan512    & \numprint{74752} \\
		 	whitaker3   & \numprint{9800} &fe\_tooth*  & \numprint{78136} \\
		 	crack       & \numprint{10240}&fe\_rotor   & \numprint{99617} \\
		 	\cline{3-4}
		 	wing\_nodal & \numprint{10937}&\multicolumn{2}{c|}{UF Graphs}\\
		 	\cline{3-4}
		 	fe\_4elt2   & \numprint{11143}& cop20k\_A       & \numprint{99843} \\
		 	vibrobox*   & \numprint{12328}& 2cubes\_sphere* & \numprint{101492}\\
		 	bcsstk29    & \numprint{13992}& thermomech\_TC  & \numprint{102158} \\
		 	4elt        & \numprint{15606}& cfd2            & \numprint{123440} \\
		 	fe\_sphere  & \numprint{16386}& boneS01         & \numprint{127224} \\
		 	cti         & \numprint{16840}& Dubcova3        & \numprint{146689} \\
		 	memplus     & \numprint{17758}& bmwcra\_1       & \numprint{148770} \\
		 	cs4         & \numprint{22499}& G2\_circuit*    & \numprint{150102} \\
		 	bcsstk30    & \numprint{28924}& c-73            & \numprint{169422} \\
		 	bcsstk31    & \numprint{35588}& shipsec5        & \numprint{179860} \\
		 	fe\_pwt     & \numprint{36519}& cont-300        & \numprint{180895} \\

		 	\cline{3-4}
		 	\hline
	\end{tabular}
        \vspace*{.25cm}
 	\caption{Walshaw graphs and florida sparse matrix graphs from \cite{2waynodeseparators}. Basic properties of the instances. Graphs with a * have been used for parameter tuning and are excluded from the evaluation.}
        \vspace*{-.5cm}
 	\label{tab:test_instances_walshaw}
\end{table}

\label{s:experiments}
\paragraph*{Methodology.} 
\ifDoubleBlind
\else
We have implemented the algorithm described above within the KaHIP framework using C++ and compiled all algorithms using gcc 4.8.3 with full optimization's turned on (-O3 flag). 
\fi{}
\ifDoubleBlind
\else
We integrated our algorithms in KaHIP v1.00. 
Our new codes will also be included into the KaHIP graph partitioning framework.
\fi{}
Each run was made on a machine that has four Octa-Core Intel Xeon E5-2670 processors running at 2.6\,GHz with 64 GB local memory.
Our main objective is the cardinality of node separators on the input graph. 
In our experiments, we use the imbalance parameter $\epsilon=3\%$ since this is one of the default values in the Metis graph partitioning framework. 
Our full algorithm is not too sensitive about the precise choice with most of the parameters.
However, we performed a number of experiments to evaluate the influence and choose the parameters of our algorithms. 
\ifDoubleBlind
\else
We omit details here and refer the reader to \cite{baWilliger}.
\fi{}
We mark the instances that have been used for the parameter tuning in Table~\ref{tab:test_instances_walshaw} with a * and exclude these graphs from our experiments.

We present multiple views on the data: average values (geometric mean) as well as convergence plots that show  quality achieved by the algorithms over time and performance plots. 
We now explain how we compute the \emph{convergence plots}.
We start explaining how we compute them for a single instance~$I$:
whenever a PE creates a separator it reports a pair ($t$, separator size), where the timestamp $t$ is the currently elapsed time on the particular PE and separator size refers to the size of the separator that has been created.
When performing multiple repetitions, we report average values ($\overline{t}$, avg. separator size) instead.
After the completion of algorithm we are left with $P$ sequences of pairs ($t$, separator size) which we now merge into one sequence.
The merged sequence is sorted by the timestamp $t$. 
\begin{figure*}
\centering
\includegraphics[width=6.5cm]{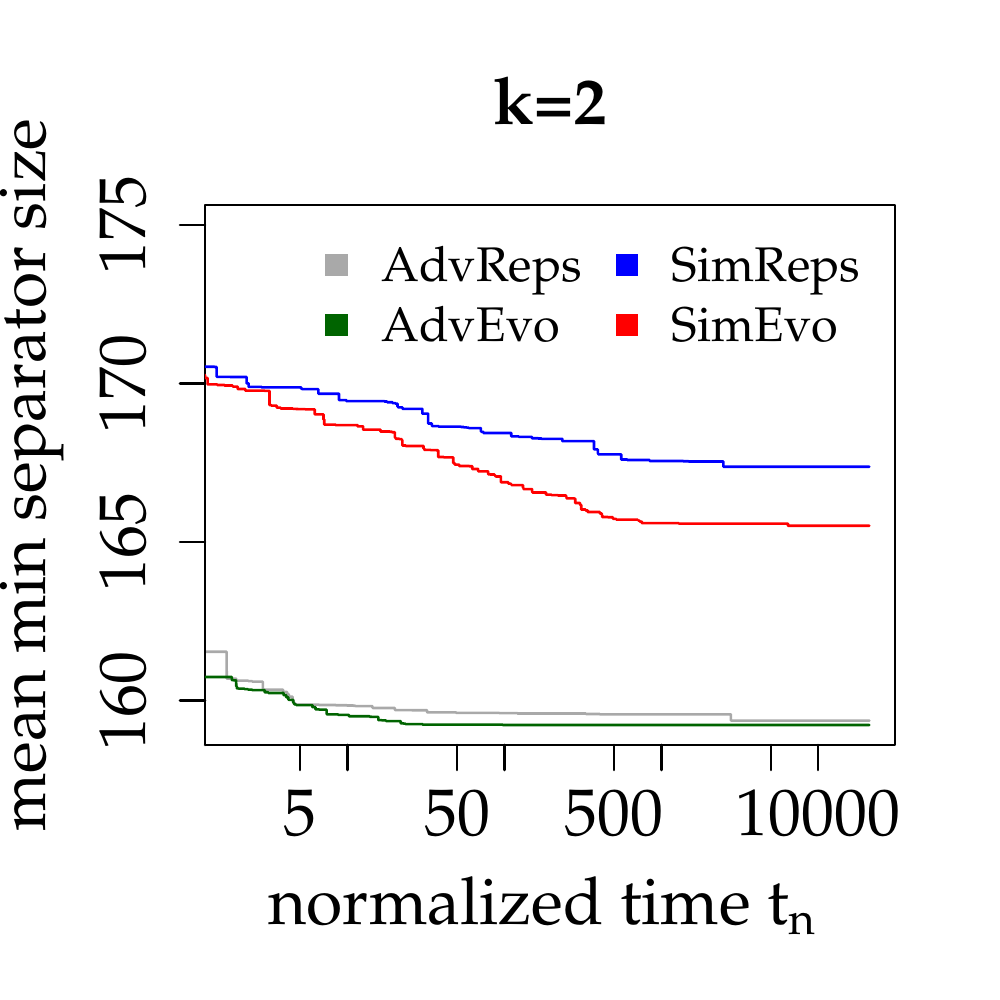}
\includegraphics[width=6.5cm]{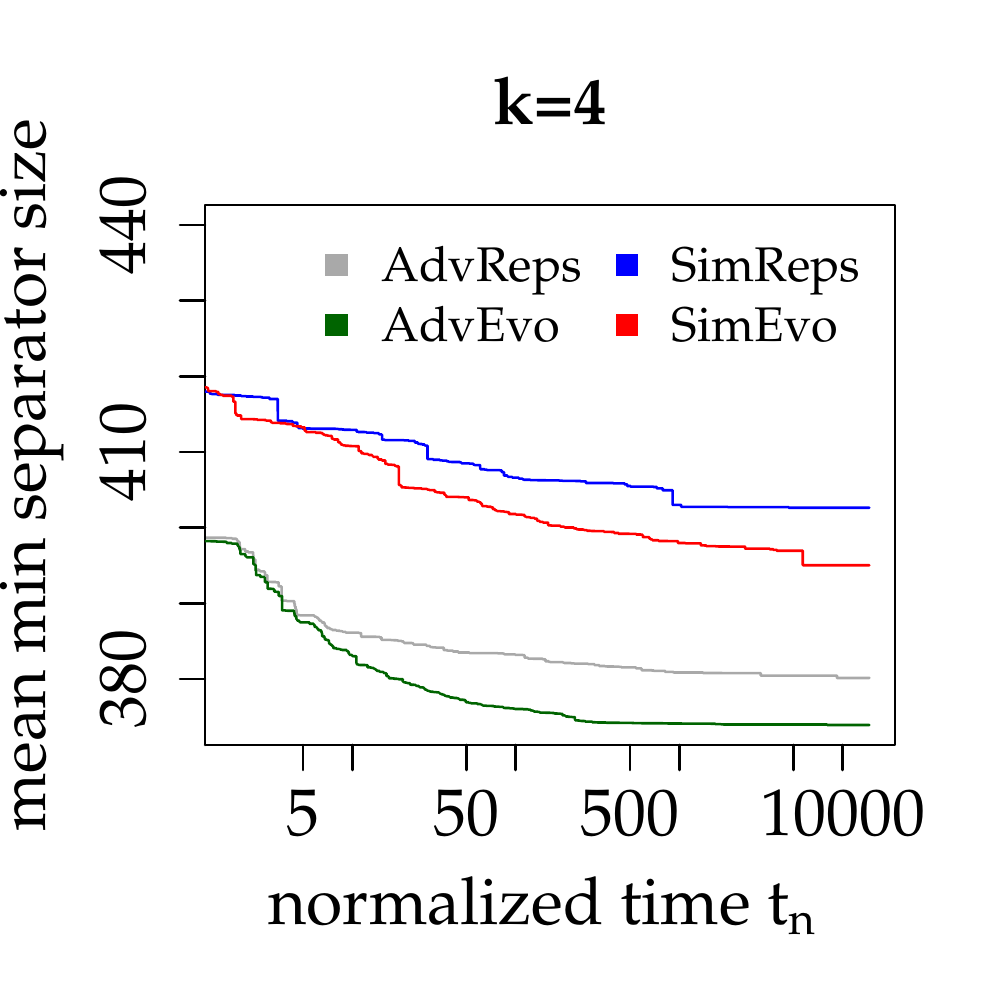} \\
\includegraphics[width=6.5cm]{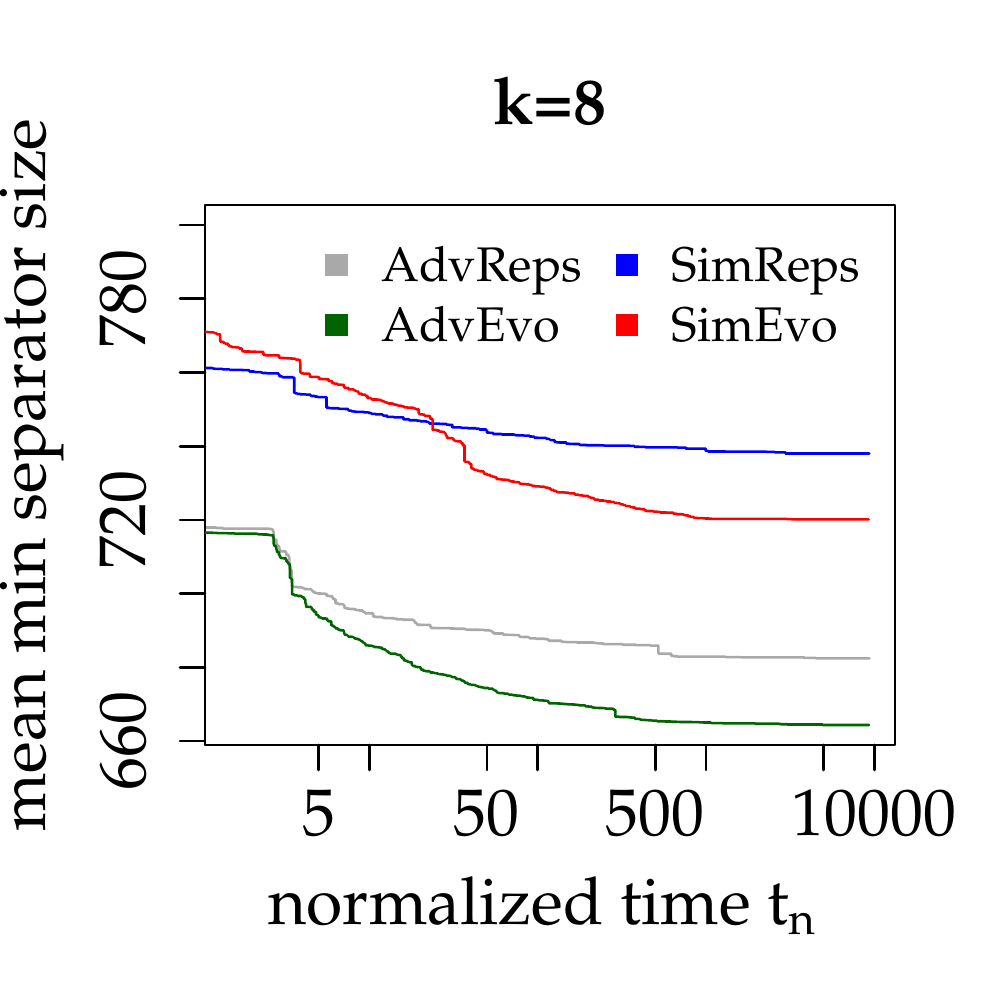} 
\includegraphics[width=6.5cm]{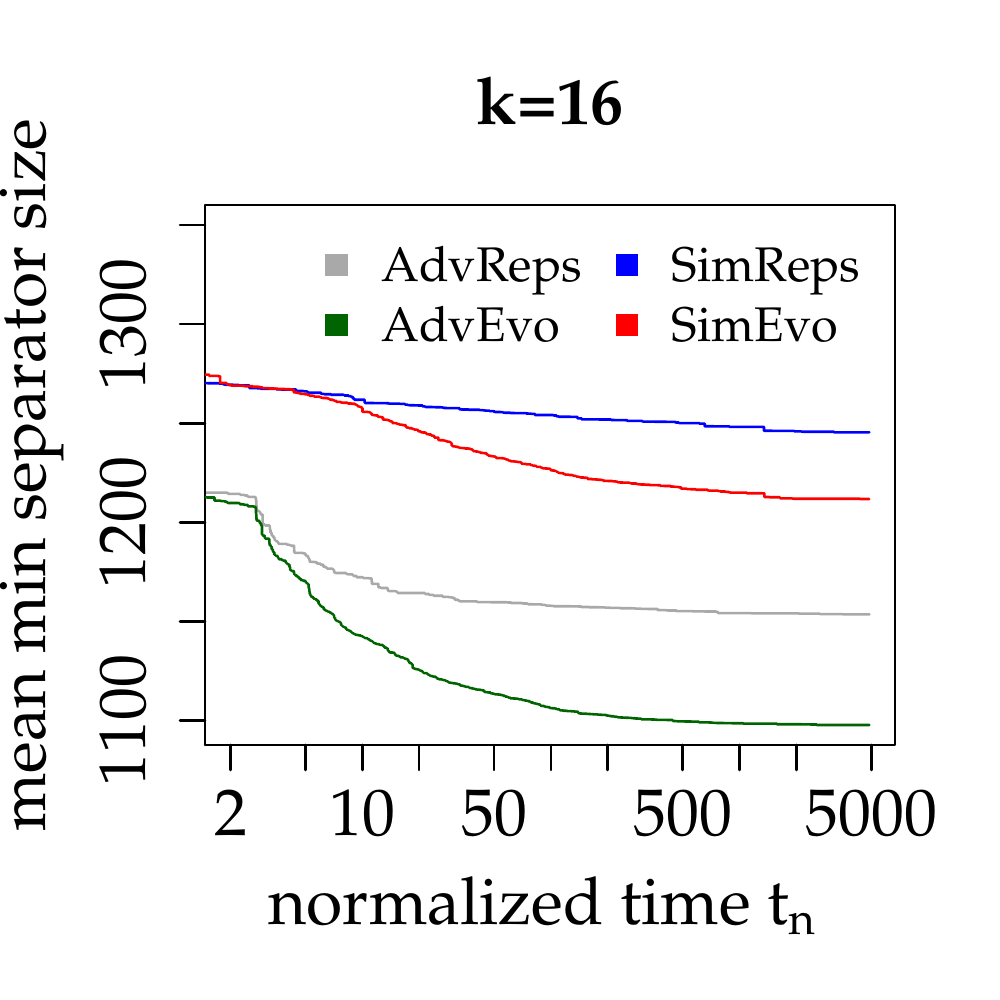} \\
\includegraphics[width=6.5cm]{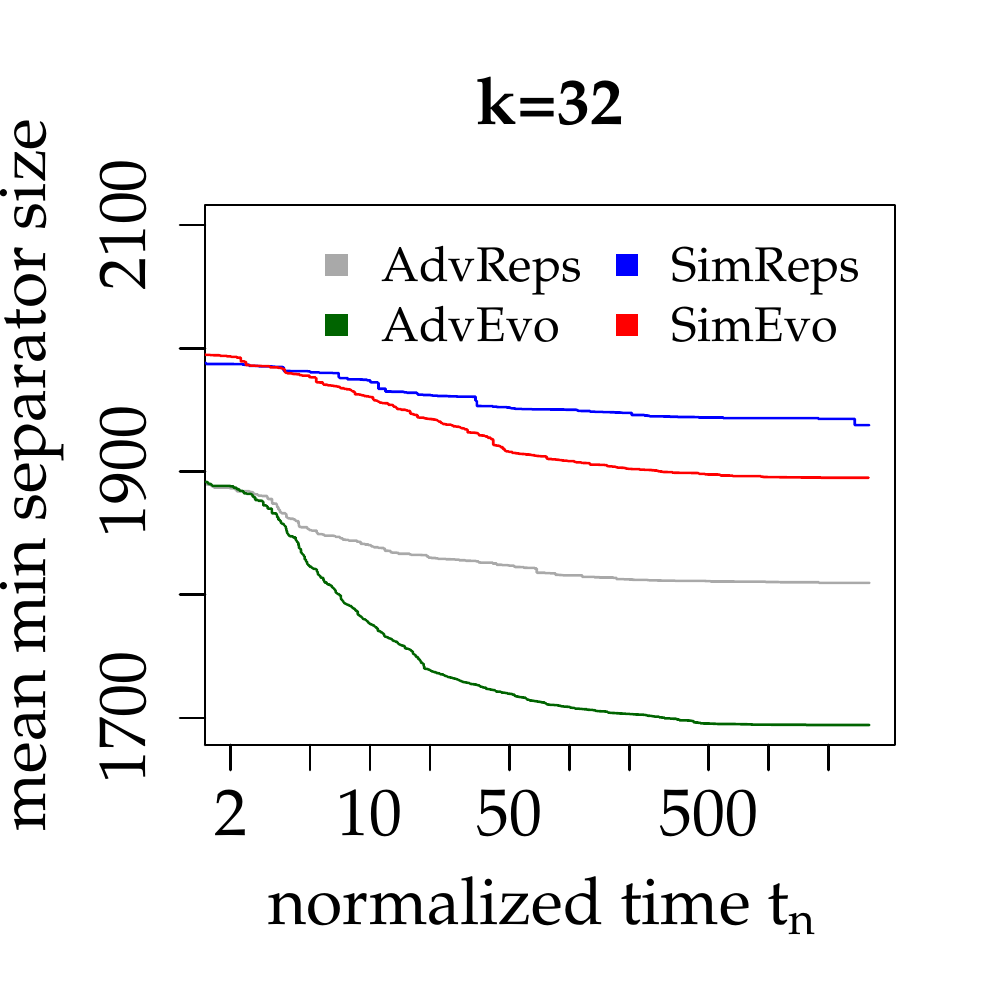} 
\includegraphics[width=6.5cm]{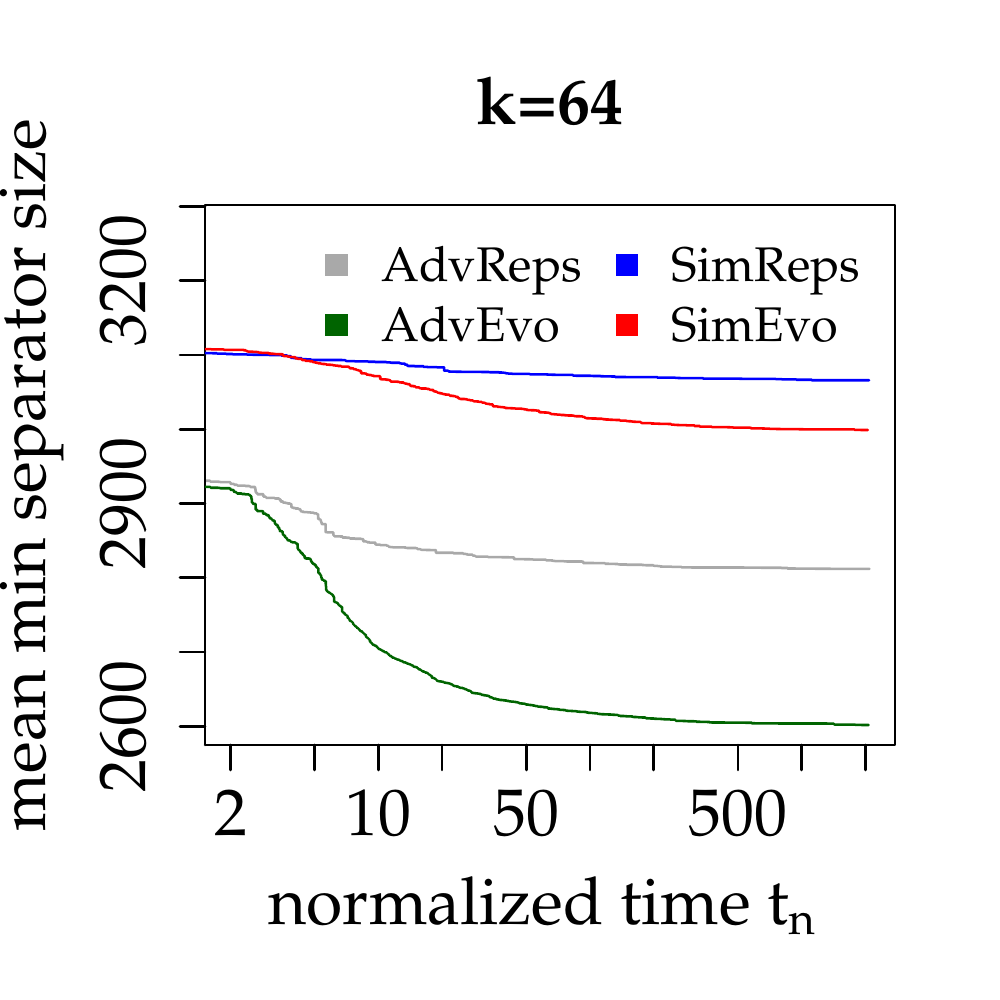} \\
\caption{Convergence plots for different values of $k$ for different algorithms.}
\label{fig:convergence_plots}
\end{figure*}
The resulting sequence is called $T^I$.
Since we are interested in the evolution of the solution quality, we compute another sequence $T^I_{\text{min}}$.
For each entry (in sorted order) in $T^I$, we insert the entry $(t, \min_{t'\leq t} \text{separator size}(t'))$ into $T^I_\text{min}$.
Here, $\min_{t'\leq t} \text{separator size}(t')$ is the minimum separator size that occurred until time $t$.
$N^I_{\text{min}}$ refers to the normalized sequence, \ie each entry ($t$, separator size) in $T^I_\text{min}$ is replaced by ($t_n$, separator size) where $t_n = t/t_I$ and $t_I$ is the average time that the sequential algorithm needs to compute a separator for the instance $I$.
To obtain average values over \emph{multiple instances} we do the following: for each instance we label all entries in $N^I_{\text{min}}$, \ie ($t_n$, separator size) is replaced by ($t_n$, separator size, $I$). We then merge all sequences $N^I_\text{min}$ and sort by $t_n$. The resulting sequence is called $S$. 
The final sequence $S_g$ presents \emph{event based} geometric averages values. 
We start by computing the geometric mean value $\mathcal{G}$ using the first value of all $N^I_\text{min}$ (over $I$).
To obtain $S_g$, we basically sweep through $S$: for each entry (in sorted order) $(t_n, c, I)$ in $S$, we update $\mathcal{G}$, \ie the separator size of $I$ that took part in the computation of $\mathcal{G}$ is replaced by the new value $c$, and insert $(t_n, \mathcal{G})$ into $S_g$. 
Note,~$c$ can be only smaller than or equal to the old value of $I$.
\paragraph*{Instances.}
We use the small and Florida Sparse Matrix graphs from \cite{2waynodeseparators} which are from various sources to test our algorithm. 
Small graphs have been obtained from Chris Walshaw's benchmark archive~\cite{soper2004combined}.
Graphs derived from sparse matrices have been taken from the Florida Sparse Matrix Collection~\cite{UFsparsematrixcollection}. 
Basic properties of the instances can be found in Table~\ref{tab:test_instances_walshaw}.

\subsection{Separator Quality}

In this section we compare our algorithms in a setting where each one gets the same (fairly large) amount of time to compute a separator. 
We do this on the graphs from our benchmark set.
We use all 16 cores per run of our machine (basically one node of the cluster) and two hours of time per instance when we use the evolutionary algorithm to create separators.
We parallelized repeated executions of the sequential algorithms (embarrassingly parallel, different seeds) and also gave them 16 PEs and two hours of time to compute a separator. 
We look at $k \in \{2,4,8,16,32,64\}$ and performed~three~repetitions~per~instance. 
To see how the solution quality of the different algorithms evolves over time, we use convergence plots. 
Figure~\ref{fig:convergence_plots} shows convergence plots for $k\in \{2,4,8,16,32,64\}$. 
Additionally, Tables~\ref{tab:detailedlongtime} and \ref{tab:numinstancesimproved} summarize final results. Tables in the Appendix present detailed per instance results.

First of all, the improvements of the evolutionary algorithms relative to repeated executions increase with growing $k$. This is due to the fact that the problems become more difficult when increasing the number of blocks $k$.
For larger values of $k$, the quality gap between the evolutionary algorithm AdvEvo and SimEvo as well as the other algorithms increases with more time invested. 
On the other hand, for $k=2$ there is almost no difference between the results produced by the evolutionary algorithm AdvEvo and the non-evolutionary version AdvReps. 
Overall, the experimental data indicates the AdvEvo is the best algorithm. Separators produced by AdvEvo are 4.1\%, 9.2\% and 11.6\% smaller compared to AdvReps, SimEvo, and SimReps on average.  
Additionally, our advanced evolutionary algorithm computes the best result on 181 out of 192~instances.

\begin{table}[t]
\small
\centering
\begin{tabular}{l||r|r|r|r}
$k$ & AdvEvo & AdvReps & SimEvo & SimReps \\
\hline
2 & 159.4 &\numprint{0.0} &+\numprint{3.8}\% &+\numprint{5.0}\% \\
4 & 373.9 &+\numprint{1.7}\%&+\numprint{5.6}\% &+\numprint{7.7}\%  \\
8 & 664.4 &+\numprint{2.7}\%&+\numprint{8.4}\% &+\numprint{11.1}\% \\
16&1097.9 &+\numprint{5.1}\%&+\numprint{10.4}\%&+\numprint{13.4}\%\\
32&1694.2 &+\numprint{6.8}\%&+\numprint{11.8}\%&+\numprint{14.4}\% \\
64&2601.8 &+\numprint{8.1}\%&+\numprint{15.3}\%&+\numprint{17.8}\% \\
        \hline
overall & +0\% & 4.1\% & 9.2\% & 11.6\% \\
\end{tabular}
\centering
\caption{Average of AdvEvo and average increase in separator size for different algorithms.}
\vspace*{-.75cm}
\label{tab:detailedlongtime}
\end{table}
Note that single executions of the simple algorithms are much faster. 
However, the results of our
 experiments 
performed in this section emphasize that one cannot simply 
\begin{table}
\centering
\small
\begin{tabular}{l||rr}
Algorithm & \#$\leq$ & \#$<$\\
\hline
AdvEvo & 181 & 122 \\
AdvReps& 65 & 7 \\
SimEvo & 33 & 3 \\
SimReps& 29 & 0 \\ 
\end{tabular}
\vspace*{.15cm}
\caption{Number of instances where algorithm X is best w.r.t to $\leq$ and $<$. The total number of instances is 192.}
\vspace*{-.75cm}
\label{tab:numinstancesimproved}
\end{table}
 use the best result out of multiple repetitions of a faster algorithm to obtain the 
same solution quality. Yet it is interesting to see that SimpleEvo, where only the fitness function of the evolutionary algorithm is modified and the combine operation still optimizes for edge cuts of partitions, computes better solutions than its non-evolutionary counter part~SimReps.

\section{Conclusion}
\label{s:conclusion}
In this work, we derived a new approach to find small node separators in large graphs which combines an evolutionary search algorithm with a multilevel method. 
We combined these techniques with a scalable communication protocol and obtain a system that is able to compute high quality solutions in a short amount of time.
Experiments show that our advanced evolutionary algorithm computes the best result on 94\% of the benchmark instances. 
In future work, we aim to look at different types of applications, in particular those applications in which the running time may not be considered a drawback when the node separator has the highest~quality.
\vspace*{-.25cm}
\paragraph*{Acknowledgements.}
The authors acknowledge support by the state of Baden-W\"urttemberg through bwHPC.
\vspace*{-.25cm}
\bibliographystyle{ACM-Reference-Format}
\bibliography{phdthesiscs} 

\begin{appendix}
\vfill
\pagebreak

\begin{table*}[!ht]
\centering
\begin{tabular}{lr||r|r||r|r||r|r||r|r|}
graph & $k$ & \multicolumn{2}{c||}{AdvEvo} & \multicolumn{2}{c||}{AdvReps} & \multicolumn{2}{c||}{SimpleEvo} & \multicolumn{2}{c|}{SimpleReps}\\
              && min & avg & min & avg & min & avg & min & avg \\
\hline
3elt &2 &\numprint{43}& \numprint{43}& \numprint{43} &\numprint{43} & \numprint{43} &\numprint{43} &\numprint{43}& \numprint{43} \\
3elt &4 &\numprint{97}& \numprint{97}& \numprint{97} &\numprint{97} & \numprint{97} &\numprint{97} &\numprint{97}& \numprint{97} \\
3elt &8 &\numprint{158}& \numprint{158}& \numprint{159} &\numprint{159} & \numprint{159} &\numprint{159} &\numprint{160}& \numprint{160} \\
3elt &16 &\numprint{269}& \numprint{269}& \numprint{270} &\numprint{270} & \numprint{271} &\numprint{271} &\numprint{272}& \numprint{272} \\
3elt &32 &\numprint{452}& \numprint{452}& \numprint{455} &\numprint{456} & \numprint{458} &\numprint{459} &\numprint{466}& \numprint{466} \\
3elt &64 &\numprint{706}& \numprint{710}& \numprint{720} &\numprint{721} & \numprint{733} &\numprint{734} &\numprint{746}& \numprint{747} \\
     \hline
4elt &2 &\numprint{68}& \numprint{68}& \numprint{68} &\numprint{68} & \numprint{68} &\numprint{68} &\numprint{68}& \numprint{68} \\
4elt &4 &\numprint{157}& \numprint{157}& \numprint{157} &\numprint{157} & \numprint{157} &\numprint{157} &\numprint{157}& \numprint{157} \\
4elt &8 &\numprint{253}& \numprint{253}& \numprint{254} &\numprint{254} & \numprint{256} &\numprint{256} &\numprint{254}& \numprint{255} \\
4elt &16 &\numprint{438}& \numprint{439}& \numprint{443} &\numprint{445} & \numprint{443} &\numprint{445} &\numprint{442}& \numprint{444} \\
4elt &32 &\numprint{737}& \numprint{743}& \numprint{755} &\numprint{757} & \numprint{744} &\numprint{746} &\numprint{749}& \numprint{750} \\
4elt &64 &\numprint{1221}& \numprint{1227}& \numprint{1257} &\numprint{1259} & \numprint{1234} &\numprint{1238} &\numprint{1253}& \numprint{1253} \\
     \hline
add20 &2 &\numprint{26}& \numprint{26}& \numprint{25} &\numprint{25} & \numprint{28} &\numprint{28} &\numprint{28}& \numprint{28} \\
add20 &4 &\numprint{37}& \numprint{37}& \numprint{37} &\numprint{37} & \numprint{45} &\numprint{47} &\numprint{49}& \numprint{49} \\
add20 &8 &\numprint{67}& \numprint{67}& \numprint{69} &\numprint{70} & \numprint{82} &\numprint{86} &\numprint{94}& \numprint{94} \\
add20 &16 &\numprint{95}& \numprint{96}& \numprint{106} &\numprint{108} & \numprint{116} &\numprint{119} &\numprint{133}& \numprint{133} \\
add20 &32 &\numprint{110}& \numprint{111}& \numprint{140} &\numprint{140} & \numprint{166} &\numprint{169} &\numprint{170}& \numprint{174} \\
add20 &64 &\numprint{138}& \numprint{138}& \numprint{161} &\numprint{164} & \numprint{219} &\numprint{225} &\numprint{218}& \numprint{221} \\
     \hline
add32 &2 &\numprint{2}& \numprint{2}& \numprint{2} &\numprint{2} & \numprint{2} &\numprint{2} &\numprint{2}& \numprint{2} \\
add32 &4 &\numprint{6}& \numprint{6}& \numprint{6} &\numprint{6} & \numprint{6} &\numprint{6} &\numprint{6}& \numprint{6} \\
add32 &8 &\numprint{11}& \numprint{11}& \numprint{11} &\numprint{11} & \numprint{12} &\numprint{12} &\numprint{12}& \numprint{12} \\
add32 &16 &\numprint{20}& \numprint{20}& \numprint{20} &\numprint{20} & \numprint{20} &\numprint{20} &\numprint{20}& \numprint{20} \\
add32 &32 &\numprint{33}& \numprint{33}& \numprint{33} &\numprint{33} & \numprint{33} &\numprint{33} &\numprint{33}& \numprint{33} \\
add32 &64 &\numprint{114}& \numprint{115}& \numprint{118} &\numprint{119} & \numprint{127} &\numprint{128} &\numprint{131}& \numprint{131} \\
     \hline
bcsstk29 &2 &\numprint{180}& \numprint{180}& \numprint{180} &\numprint{180} & \numprint{180} &\numprint{180} &\numprint{180}& \numprint{180} \\
bcsstk29 &4 &\numprint{528}& \numprint{528}& \numprint{528} &\numprint{528} & \numprint{534} &\numprint{536} &\numprint{534}& \numprint{534} \\
bcsstk29 &8 &\numprint{948}& \numprint{954}& \numprint{966} &\numprint{968} & \numprint{1173} &\numprint{1202} &\numprint{1086}& \numprint{1094} \\
bcsstk29 &16 &\numprint{1512}& \numprint{1530}& \numprint{1578} &\numprint{1578} & \numprint{2120} &\numprint{2127} &\numprint{2019}& \numprint{2062} \\
bcsstk29 &32 &\numprint{2231}& \numprint{2250}& \numprint{2316} &\numprint{2326} & \numprint{2891} &\numprint{2898} &\numprint{2892}& \numprint{2899} \\
bcsstk29 &64 &\numprint{3130}& \numprint{3157}& \numprint{3361} &\numprint{3371} & \numprint{4073} &\numprint{4089} &\numprint{4065}& \numprint{4096} \\
     \hline
bcsstk30 &2 &\numprint{206}& \numprint{206}& \numprint{206} &\numprint{206} & \numprint{206} &\numprint{206} &\numprint{206}& \numprint{206} \\
bcsstk30 &4 &\numprint{549}& \numprint{549}& \numprint{549} &\numprint{549} & \numprint{573} &\numprint{573} &\numprint{573}& \numprint{573} \\
bcsstk30 &8 &\numprint{1121}& \numprint{1121}& \numprint{1123} &\numprint{1123} & \numprint{1138} &\numprint{1145} &\numprint{1138}& \numprint{1140} \\
bcsstk30 &16 &\numprint{2128}& \numprint{2146}& \numprint{2183} &\numprint{2201} & \numprint{2446} &\numprint{2455} &\numprint{2430}& \numprint{2452} \\
bcsstk30 &32 &\numprint{3195}& \numprint{3249}& \numprint{3321} &\numprint{3338} & \numprint{3985} &\numprint{3987} &\numprint{3956}& \numprint{4008} \\
bcsstk30 &64 &\numprint{4709}& \numprint{4836}& \numprint{5045} &\numprint{5111} & \numprint{5846} &\numprint{5940} &\numprint{5994}& \numprint{6013} \\
     \hline
bcsstk31 &2 &\numprint{308}& \numprint{308}& \numprint{308} &\numprint{308} & \numprint{317} &\numprint{321} &\numprint{317}& \numprint{319} \\
bcsstk31 &4 &\numprint{767}& \numprint{767}& \numprint{767} &\numprint{767} & \numprint{802} &\numprint{803} &\numprint{798}& \numprint{800} \\
bcsstk31 &8 &\numprint{1433}& \numprint{1434}& \numprint{1441} &\numprint{1442} & \numprint{1529} &\numprint{1538} &\numprint{1534}& \numprint{1545} \\
bcsstk31 &16 &\numprint{2353}& \numprint{2399}& \numprint{2437} &\numprint{2446} & \numprint{2592} &\numprint{2630} &\numprint{2624}& \numprint{2633} \\
bcsstk31 &32 &\numprint{3635}& \numprint{3695}& \numprint{3837} &\numprint{3874} & \numprint{4338} &\numprint{4361} &\numprint{4361}& \numprint{4376} \\
bcsstk31 &64 &\numprint{5102}& \numprint{5167}& \numprint{5323} &\numprint{5394} & \numprint{6090} &\numprint{6136} &\numprint{6169}& \numprint{6205} \\
     \hline
bcsstk32 &2 &\numprint{297}& \numprint{297}& \numprint{297} &\numprint{297} & \numprint{322} &\numprint{328} &\numprint{321}& \numprint{321} \\
bcsstk32 &4 &\numprint{569}& \numprint{569}& \numprint{569} &\numprint{569} & \numprint{633} &\numprint{637} &\numprint{627}& \numprint{627} \\
bcsstk32 &8 &\numprint{1145}& \numprint{1152}& \numprint{1177} &\numprint{1183} & \numprint{1312} &\numprint{1336} &\numprint{1315}& \numprint{1326} \\
bcsstk32 &16 &\numprint{2080}& \numprint{2102}& \numprint{2122} &\numprint{2131} & \numprint{2391} &\numprint{2434} &\numprint{2443}& \numprint{2468} \\
bcsstk32 &32 &\numprint{3422}& \numprint{3449}& \numprint{3498} &\numprint{3524} & \numprint{4102} &\numprint{4118} &\numprint{4114}& \numprint{4142} \\
bcsstk32 &64 &\numprint{5386}& \numprint{5469}& \numprint{5621} &\numprint{5677} & \numprint{6293} &\numprint{6321} &\numprint{6348}& \numprint{6412} \\
     \hline
bmwcra\_1 &2 &\numprint{657}& \numprint{657}& \numprint{657} &\numprint{657} & \numprint{684} &\numprint{684} &\numprint{683}& \numprint{683} \\
bmwcra\_1 &4 &\numprint{1668}& \numprint{1673}& \numprint{1656} &\numprint{1659} & \numprint{1683} &\numprint{1685} &\numprint{1659}& \numprint{1666} \\
bmwcra\_1 &8 &\numprint{3918}& \numprint{3970}& \numprint{4002} &\numprint{4013} & \numprint{4080} &\numprint{4126} &\numprint{4112}& \numprint{4133} \\
bmwcra\_1 &16 &\numprint{10011}& \numprint{10037}& \numprint{9846} &\numprint{9921} & \numprint{10099} &\numprint{10199} &\numprint{10190}& \numprint{10250} \\
bmwcra\_1 &32 &\numprint{16089}& \numprint{16300}& \numprint{16798} &\numprint{16979} & \numprint{16863} &\numprint{16947} &\numprint{16725}& \numprint{16775} \\
bmwcra\_1 &64 &\numprint{23586}& \numprint{24279}& \numprint{25707} &\numprint{25809} & \numprint{24979} &\numprint{25087} &\numprint{24885}& \numprint{25042} \\
\hline
\end{tabular}
\caption{Detailed per instance results.}
\label{tab:detailedlongtimeone}
\end{table*}
\FloatBarrier
\begin{table*}\centering\begin{tabular}{lr||r|r||r|r||r|r||r|r|}
graph & $k$ & \multicolumn{2}{c||}{AdvEvo} & \multicolumn{2}{c||}{AdvReps} & \multicolumn{2}{c||}{SimpleEvo} & \multicolumn{2}{c|}{SimpleReps}\\
              && min & avg & min & avg & min & avg & min & avg \\
\hline

boneS01 &2 &\numprint{1524}& \numprint{1524}& \numprint{1524} &\numprint{1524} & \numprint{1563} &\numprint{1563} &\numprint{1563}& \numprint{1565} \\
boneS01 &4 &\numprint{3357}& \numprint{3357}& \numprint{3357} &\numprint{3357} & \numprint{3465} &\numprint{3469} &\numprint{3471}& \numprint{3477} \\
boneS01 &8 &\numprint{5112}& \numprint{5128}& \numprint{5139} &\numprint{5148} & \numprint{5316} &\numprint{5351} &\numprint{5358}& \numprint{5371} \\
boneS01 &16 &\numprint{7728}& \numprint{7781}& \numprint{7776} &\numprint{7826} & \numprint{8139} &\numprint{8179} &\numprint{8142}& \numprint{8189} \\
boneS01 &32 &\numprint{11082}& \numprint{11127}& \numprint{11400} &\numprint{11440} & \numprint{11619} &\numprint{11670} &\numprint{11758}& \numprint{11770} \\
boneS01 &64 &\numprint{15264}& \numprint{15271}& \numprint{16495} &\numprint{16683} & \numprint{16173} &\numprint{16219} &\numprint{16296}& \numprint{16314} \\
        \hline
brack2 &2 &\numprint{183}& \numprint{183}& \numprint{183} &\numprint{183} & \numprint{206} &\numprint{208} &\numprint{205}& \numprint{205} \\
brack2 &4 &\numprint{796}& \numprint{796}& \numprint{796} &\numprint{796} & \numprint{829} &\numprint{831} &\numprint{829}& \numprint{829} \\
brack2 &8 &\numprint{1740}& \numprint{1741}& \numprint{1759} &\numprint{1761} & \numprint{1896} &\numprint{1899} &\numprint{1906}& \numprint{1910} \\
brack2 &16 &\numprint{2794}& \numprint{2806}& \numprint{2868} &\numprint{2874} & \numprint{3108} &\numprint{3119} &\numprint{3117}& \numprint{3120} \\
brack2 &32 &\numprint{4160}& \numprint{4184}& \numprint{4282} &\numprint{4289} & \numprint{4632} &\numprint{4647} &\numprint{4646}& \numprint{4651} \\
brack2 &64 &\numprint{6053}& \numprint{6084}& \numprint{6361} &\numprint{6381} & \numprint{6808} &\numprint{6841} &\numprint{6809}& \numprint{6820} \\
        \hline
cfd2 &2 &\numprint{1030}& \numprint{1030}& \numprint{1030} &\numprint{1030} & \numprint{1036} &\numprint{1036} &\numprint{1036}& \numprint{1036} \\
cfd2 &4 &\numprint{2543}& \numprint{2546}& \numprint{2548} &\numprint{2551} & \numprint{2684} &\numprint{2688} &\numprint{2645}& \numprint{2667} \\
cfd2 &8 &\numprint{4304}& \numprint{4313}& \numprint{4304} &\numprint{4312} & \numprint{4569} &\numprint{4581} &\numprint{4551}& \numprint{4591} \\
cfd2 &16 &\numprint{7068}& \numprint{7095}& \numprint{7018} &\numprint{7036} & \numprint{7416} &\numprint{7440} &\numprint{7374}& \numprint{7392} \\
cfd2 &32 &\numprint{10723}& \numprint{10873}& \numprint{11165} &\numprint{11272} & \numprint{12066} &\numprint{12088} &\numprint{11924}& \numprint{11956} \\
cfd2 &64 &\numprint{16521}& \numprint{17138}& \numprint{17829} &\numprint{18021} & \numprint{18093} &\numprint{18179} &\numprint{18014}& \numprint{18067} \\
\hline
cont-300 &2 &\numprint{598}& \numprint{598}& \numprint{598} &\numprint{598} & \numprint{598} &\numprint{598} &\numprint{598}& \numprint{598} \\
cont-300 &4 &\numprint{1041}& \numprint{1042}& \numprint{1063} &\numprint{1065} & \numprint{1184} &\numprint{1184} &\numprint{1184}& \numprint{1184} \\
cont-300 &8 &\numprint{1786}& \numprint{1814}& \numprint{1807} &\numprint{1825} & \numprint{2188} &\numprint{2192} &\numprint{2188}& \numprint{2194} \\
cont-300 &16 &\numprint{2863}& \numprint{2874}& \numprint{2893} &\numprint{2916} & \numprint{3526} &\numprint{3528} &\numprint{3534}& \numprint{3537} \\
cont-300 &32 &\numprint{4299}& \numprint{4340}& \numprint{4413} &\numprint{4450} & \numprint{5466} &\numprint{5483} &\numprint{5504}& \numprint{5506} \\
cont-300 &64 &\numprint{6452}& \numprint{6474}& \numprint{6667} &\numprint{6679} & \numprint{8094} &\numprint{8105} &\numprint{8168}& \numprint{8170} \\
\hline
cop20k\_A &2 &\numprint{620}& \numprint{620}& \numprint{620} &\numprint{620} & \numprint{620} &\numprint{620} &\numprint{620}& \numprint{620} \\
cop20k\_A &4 &\numprint{1673}& \numprint{1675}& \numprint{1676} &\numprint{1676} & \numprint{1733} &\numprint{1741} &\numprint{1716}& \numprint{1724} \\
cop20k\_A &8 &\numprint{2919}& \numprint{2934}& \numprint{2939} &\numprint{2942} & \numprint{2997} &\numprint{2999} &\numprint{2993}& \numprint{2996} \\
cop20k\_A &16 &\numprint{4721}& \numprint{4744}& \numprint{4765} &\numprint{4780} & \numprint{4842} &\numprint{4864} &\numprint{4849}& \numprint{4858} \\
cop20k\_A &32 &\numprint{7241}& \numprint{7333}& \numprint{7525} &\numprint{7645} & \numprint{7481} &\numprint{7502} &\numprint{7423}& \numprint{7465} \\
cop20k\_A &64 &\numprint{10757}& \numprint{10837}& \numprint{11721} &\numprint{12107} & \numprint{11135} &\numprint{11147} &\numprint{11102}& \numprint{11155} \\
\hline
crack &2 &\numprint{73}& \numprint{73}& \numprint{73} &\numprint{73} & \numprint{75} &\numprint{75} &\numprint{74}& \numprint{74} \\
crack &4 &\numprint{145}& \numprint{145}& \numprint{145} &\numprint{145} & \numprint{152} &\numprint{152} &\numprint{152}& \numprint{152} \\
crack &8 &\numprint{257}& \numprint{257}& \numprint{258} &\numprint{258} & \numprint{280} &\numprint{282} &\numprint{285}& \numprint{286} \\
crack &16 &\numprint{420}& \numprint{420}& \numprint{421} &\numprint{422} & \numprint{461} &\numprint{465} &\numprint{474}& \numprint{474} \\
crack &32 &\numprint{636}& \numprint{639}& \numprint{648} &\numprint{648} & \numprint{722} &\numprint{724} &\numprint{735}& \numprint{738} \\
crack &64 &\numprint{939}& \numprint{943}& \numprint{958} &\numprint{959} & \numprint{1098} &\numprint{1100} &\numprint{1123}& \numprint{1124} \\
\hline
cs4 &2 &\numprint{287}& \numprint{287}& \numprint{287} &\numprint{287} & \numprint{319} &\numprint{319} &\numprint{322}& \numprint{322} \\
cs4 &4 &\numprint{727}& \numprint{729}& \numprint{738} &\numprint{740} & \numprint{824} &\numprint{825} &\numprint{832}& \numprint{836} \\
cs4 &8 &\numprint{1108}& \numprint{1109}& \numprint{1133} &\numprint{1134} & \numprint{1244} &\numprint{1245} &\numprint{1267}& \numprint{1267} \\
cs4 &16 &\numprint{1548}& \numprint{1558}& \numprint{1623} &\numprint{1630} & \numprint{1812} &\numprint{1818} &\numprint{1827}& \numprint{1830} \\
cs4 &32 &\numprint{2132}& \numprint{2148}& \numprint{2273} &\numprint{2286} & \numprint{2512} &\numprint{2516} &\numprint{2537}& \numprint{2541} \\
cs4 &64 &\numprint{2864}& \numprint{2909}& \numprint{3179} &\numprint{3202} & \numprint{3433} &\numprint{3435} &\numprint{3461}& \numprint{3463} \\
\hline
cti &2 &\numprint{266}& \numprint{266}& \numprint{266} &\numprint{266} & \numprint{266} &\numprint{266} &\numprint{266}& \numprint{266} \\
cti &4 &\numprint{756}& \numprint{758}& \numprint{761} &\numprint{761} & \numprint{808} &\numprint{808} &\numprint{807}& \numprint{807} \\
cti &8 &\numprint{1243}& \numprint{1270}& \numprint{1311} &\numprint{1315} & \numprint{1537} &\numprint{1539} &\numprint{1539}& \numprint{1539} \\
cti &16 &\numprint{1821}& \numprint{1845}& \numprint{1897} &\numprint{1925} & \numprint{2287} &\numprint{2298} &\numprint{2319}& \numprint{2324} \\
cti &32 &\numprint{2426}& \numprint{2457}& \numprint{2646} &\numprint{2647} & \numprint{3257} &\numprint{3267} &\numprint{3258}& \numprint{3267} \\
cti &64 &\numprint{3234}& \numprint{3242}& \numprint{3581} &\numprint{3596} & \numprint{4489} &\numprint{4491} &\numprint{4481}& \numprint{4491} \\
\hline
data &2 &\numprint{51}& \numprint{51}& \numprint{51} &\numprint{51} & \numprint{51} &\numprint{51} &\numprint{51}& \numprint{51} \\
data &4 &\numprint{96}& \numprint{96}& \numprint{96} &\numprint{96} & \numprint{96} &\numprint{96} &\numprint{97}& \numprint{97} \\
data &8 &\numprint{165}& \numprint{165}& \numprint{165} &\numprint{165} & \numprint{172} &\numprint{172} &\numprint{173}& \numprint{173} \\
data &16 &\numprint{275}& \numprint{275}& \numprint{276} &\numprint{276} & \numprint{298} &\numprint{301} &\numprint{300}& \numprint{300} \\
data &32 &\numprint{448}& \numprint{450}& \numprint{454} &\numprint{454} & \numprint{489} &\numprint{492} &\numprint{499}& \numprint{502} \\
data &64 &\numprint{688}& \numprint{690}& \numprint{695} &\numprint{697} & \numprint{757} &\numprint{760} &\numprint{764}& \numprint{764} \\
\hline
\end{tabular}
\caption{Detailed per instance results.}
\label{tab:detailedlongtimetwo}
\end{table*}
\begin{table*}\centering\begin{tabular}{lr||r|r||r|r||r|r||r|r|}
graph & $k$ & \multicolumn{2}{c||}{AdvEvo} & \multicolumn{2}{c||}{AdvReps} & \multicolumn{2}{c||}{SimpleEvo} & \multicolumn{2}{c|}{SimpleReps}\\
              && min & avg & min & avg & min & avg & min & avg \\
\hline

Dubcova3 &2 &\numprint{383}& \numprint{383}& \numprint{383} &\numprint{383} & \numprint{383} &\numprint{383} &\numprint{383}& \numprint{383} \\
Dubcova3 &4 &\numprint{765}& \numprint{765}& \numprint{765} &\numprint{765} & \numprint{765} &\numprint{765} &\numprint{765}& \numprint{765} \\
Dubcova3 &8 &\numprint{1433}& \numprint{1437}& \numprint{1436} &\numprint{1437} & \numprint{1463} &\numprint{1475} &\numprint{1463}& \numprint{1464} \\
Dubcova3 &16 &\numprint{2295}& \numprint{2309}& \numprint{2319} &\numprint{2322} & \numprint{2373} &\numprint{2398} &\numprint{2346}& \numprint{2355} \\
Dubcova3 &32 &\numprint{3581}& \numprint{3621}& \numprint{3707} &\numprint{3711} & \numprint{3887} &\numprint{3893} &\numprint{3821}& \numprint{3856} \\
Dubcova3 &64 &\numprint{5448}& \numprint{5492}& \numprint{5706} &\numprint{5732} & \numprint{5754} &\numprint{5772} &\numprint{5763}& \numprint{5766} \\
         \hline
fe\_4elt2 &2 &\numprint{66}& \numprint{66}& \numprint{66} &\numprint{66} & \numprint{66} &\numprint{66} &\numprint{66}& \numprint{66} \\
fe\_4elt2 &4 &\numprint{163}& \numprint{163}& \numprint{162} &\numprint{162} & \numprint{168} &\numprint{169} &\numprint{167}& \numprint{167} \\
fe\_4elt2 &8 &\numprint{283}& \numprint{285}& \numprint{288} &\numprint{288} & \numprint{290} &\numprint{291} &\numprint{293}& \numprint{293} \\
fe\_4elt2 &16 &\numprint{478}& \numprint{479}& \numprint{482} &\numprint{483} & \numprint{482} &\numprint{484} &\numprint{486}& \numprint{487} \\
fe\_4elt2 &32 &\numprint{759}& \numprint{762}& \numprint{780} &\numprint{781} & \numprint{773} &\numprint{774} &\numprint{783}& \numprint{785} \\
fe\_4elt2 &64 &\numprint{1149}& \numprint{1153}& \numprint{1185} &\numprint{1190} & \numprint{1189} &\numprint{1191} &\numprint{1206}& \numprint{1208} \\
         \hline
fe\_pwt &2 &\numprint{116}& \numprint{116}& \numprint{116} &\numprint{116} & \numprint{116} &\numprint{116} &\numprint{116}& \numprint{116} \\
fe\_pwt &4 &\numprint{236}& \numprint{236}& \numprint{236} &\numprint{236} & \numprint{236} &\numprint{236} &\numprint{236}& \numprint{236} \\
fe\_pwt &8 &\numprint{473}& \numprint{473}& \numprint{473} &\numprint{473} & \numprint{474} &\numprint{474} &\numprint{476}& \numprint{476} \\
fe\_pwt &16 &\numprint{925}& \numprint{925}& \numprint{929} &\numprint{929} & \numprint{930} &\numprint{930} &\numprint{929}& \numprint{929} \\
fe\_pwt &32 &\numprint{1834}& \numprint{1839}& \numprint{1909} &\numprint{1910} & \numprint{1862} &\numprint{1864} &\numprint{1872}& \numprint{1873} \\
fe\_pwt &64 &\numprint{2846}& \numprint{2919}& \numprint{3025} &\numprint{3043} & \numprint{3458} &\numprint{3472} &\numprint{3447}& \numprint{3462} \\
         \hline
fe\_rotor &2 &\numprint{460}& \numprint{460}& \numprint{460} &\numprint{460} & \numprint{464} &\numprint{464} &\numprint{464}& \numprint{464} \\
fe\_rotor &4 &\numprint{1540}& \numprint{1545}& \numprint{1543} &\numprint{1554} & \numprint{1575} &\numprint{1593} &\numprint{1573}& \numprint{1580} \\
fe\_rotor &8 &\numprint{2833}& \numprint{2838}& \numprint{2844} &\numprint{2848} & \numprint{2891} &\numprint{2891} &\numprint{2898}& \numprint{2899} \\
fe\_rotor &16 &\numprint{4404}& \numprint{4448}& \numprint{4483} &\numprint{4489} & \numprint{4605} &\numprint{4632} &\numprint{4550}& \numprint{4589} \\
fe\_rotor &32 &\numprint{6809}& \numprint{6898}& \numprint{6943} &\numprint{7013} & \numprint{7024} &\numprint{7065} &\numprint{7037}& \numprint{7062} \\
fe\_rotor &64 &\numprint{10196}& \numprint{10289}& \numprint{10534} &\numprint{10615} & \numprint{10249} &\numprint{10293} &\numprint{10242}& \numprint{10261} \\
         \hline
fe\_sphere &2 &\numprint{192}& \numprint{192}& \numprint{192} &\numprint{192} & \numprint{192} &\numprint{192} &\numprint{192}& \numprint{192} \\
fe\_sphere &4 &\numprint{379}& \numprint{379}& \numprint{380} &\numprint{380} & \numprint{380} &\numprint{380} &\numprint{380}& \numprint{380} \\
fe\_sphere &8 &\numprint{570}& \numprint{570}& \numprint{575} &\numprint{577} & \numprint{570} &\numprint{570} &\numprint{570}& \numprint{570} \\
fe\_sphere &16 &\numprint{804}& \numprint{804}& \numprint{852} &\numprint{854} & \numprint{835} &\numprint{835} &\numprint{839}& \numprint{839} \\
fe\_sphere &32 &\numprint{1177}& \numprint{1184}& \numprint{1262} &\numprint{1264} & \numprint{1208} &\numprint{1213} &\numprint{1216}& \numprint{1218} \\
fe\_sphere &64 &\numprint{1657}& \numprint{1667}& \numprint{1803} &\numprint{1805} & \numprint{1722} &\numprint{1723} &\numprint{1745}& \numprint{1750} \\
         \hline
finan512 &2 &\numprint{50}& \numprint{50}& \numprint{50} &\numprint{50} & \numprint{50} &\numprint{50} &\numprint{50}& \numprint{50} \\
finan512 &4 &\numprint{100}& \numprint{100}& \numprint{100} &\numprint{100} & \numprint{100} &\numprint{100} &\numprint{100}& \numprint{100} \\
finan512 &8 &\numprint{200}& \numprint{200}& \numprint{200} &\numprint{200} & \numprint{200} &\numprint{200} &\numprint{200}& \numprint{200} \\
finan512 &16 &\numprint{400}& \numprint{400}& \numprint{400} &\numprint{400} & \numprint{400} &\numprint{400} &\numprint{400}& \numprint{400} \\
finan512 &32 &\numprint{800}& \numprint{800}& \numprint{800} &\numprint{800} & \numprint{800} &\numprint{800} &\numprint{800}& \numprint{800} \\
finan512 &64 &\numprint{3210}& \numprint{3216}& \numprint{3259} &\numprint{3263} & \numprint{3200} &\numprint{3200} &\numprint{3200}& \numprint{3200} \\
         \hline
memplus &2 &\numprint{70}& \numprint{70}& \numprint{70} &\numprint{70} & \numprint{90} &\numprint{103} &\numprint{107}& \numprint{107} \\
memplus &4 &\numprint{90}& \numprint{90}& \numprint{91} &\numprint{91} & \numprint{127} &\numprint{131} &\numprint{123}& \numprint{126} \\
memplus &8 &\numprint{106}& \numprint{106}& \numprint{106} &\numprint{106} & \numprint{154} &\numprint{158} &\numprint{142}& \numprint{144} \\
memplus &16 &\numprint{132}& \numprint{132}& \numprint{151} &\numprint{153} & \numprint{234} &\numprint{240} &\numprint{239}& \numprint{242} \\
memplus &32 &\numprint{178}& \numprint{179}& \numprint{194} &\numprint{195} & \numprint{265} &\numprint{268} &\numprint{264}& \numprint{268} \\
memplus &64 &\numprint{181}& \numprint{182}& \numprint{205} &\numprint{206} & \numprint{392} &\numprint{412} &\numprint{428}& \numprint{436} \\
         \hline
shipsec5 &2 &\numprint{1203}& \numprint{1203}& \numprint{1203} &\numprint{1203} & \numprint{1227} &\numprint{1231} &\numprint{1227}& \numprint{1233} \\
shipsec5 &4 &\numprint{3681}& \numprint{3681}& \numprint{3681} &\numprint{3681} & \numprint{3783} &\numprint{3793} &\numprint{3801}& \numprint{3803} \\
shipsec5 &8 &\numprint{6078}& \numprint{6112}& \numprint{6198} &\numprint{6216} & \numprint{6486} &\numprint{6509} &\numprint{6531}& \numprint{6549} \\
shipsec5 &16 &\numprint{8826}& \numprint{8881}& \numprint{8850} &\numprint{8903} & \numprint{9612} &\numprint{9650} &\numprint{9570}& \numprint{9651} \\
shipsec5 &32 &\numprint{12861}& \numprint{12983}& \numprint{13521} &\numprint{13601} & \numprint{14208} &\numprint{14236} &\numprint{14316}& \numprint{14340} \\
shipsec5 &64 &\numprint{17304}& \numprint{17398}& \numprint{18114} &\numprint{18301} & \numprint{21482} &\numprint{21594} &\numprint{21549}& \numprint{21611} \\
         \hline
t60k &2 &\numprint{70}& \numprint{70}& \numprint{70} &\numprint{70} & \numprint{71} &\numprint{71} &\numprint{71}& \numprint{71} \\
t60k &4 &\numprint{202}& \numprint{202}& \numprint{202} &\numprint{202} & \numprint{203} &\numprint{203} &\numprint{203}& \numprint{203} \\
t60k &8 &\numprint{447}& \numprint{447}& \numprint{449} &\numprint{449} & \numprint{448} &\numprint{448} &\numprint{448}& \numprint{448} \\
t60k &16 &\numprint{800}& \numprint{803}& \numprint{807} &\numprint{810} & \numprint{792} &\numprint{793} &\numprint{802}& \numprint{804} \\
t60k &32 &\numprint{1330}& \numprint{1331}& \numprint{1339} &\numprint{1343} & \numprint{1307} &\numprint{1308} &\numprint{1335}& \numprint{1336} \\
t60k &64 &\numprint{2040}& \numprint{2042}& \numprint{2104} &\numprint{2114} & \numprint{2031} &\numprint{2043} &\numprint{2098}& \numprint{2103} \\
\hline
\end{tabular}
\caption{Detailed per instance results.}
\label{tab:detailedlongtimethree}
\end{table*}
\begin{table*}\centering\begin{tabular}{lr||r|r||r|r||r|r||r|r|}
graph & $k$ & \multicolumn{2}{c||}{AdvEvo} & \multicolumn{2}{c||}{AdvReps} & \multicolumn{2}{c||}{SimpleEvo} & \multicolumn{2}{c|}{SimpleReps}\\
              && min & avg & min & avg & min & avg & min & avg \\
\hline

thermomech\_TC &2 &\numprint{128}& \numprint{128}& \numprint{128} &\numprint{128} & \numprint{129} &\numprint{129} &\numprint{129}& \numprint{129} \\
thermomech\_TC &4 &\numprint{390}& \numprint{390}& \numprint{389} &\numprint{389} & \numprint{394} &\numprint{394} &\numprint{394}& \numprint{394} \\
thermomech\_TC &8 &\numprint{837}& \numprint{839}& \numprint{838} &\numprint{838} & \numprint{843} &\numprint{844} &\numprint{848}& \numprint{848} \\
thermomech\_TC &16 &\numprint{1439}& \numprint{1440}& \numprint{1447} &\numprint{1450} & \numprint{1445} &\numprint{1447} &\numprint{1453}& \numprint{1456} \\
thermomech\_TC &32 &\numprint{2316}& \numprint{2323}& \numprint{2345} &\numprint{2349} & \numprint{2340} &\numprint{2347} &\numprint{2345}& \numprint{2353} \\
thermomech\_TC &64 &\numprint{3620}& \numprint{3627}& \numprint{3672} &\numprint{3680} & \numprint{3636} &\numprint{3645} &\numprint{3659}& \numprint{3663} \\
     \hline
uk &2 &\numprint{18}& \numprint{18}& \numprint{18} &\numprint{18} & \numprint{18} &\numprint{18} &\numprint{18}& \numprint{18} \\
uk &4 &\numprint{37}& \numprint{37}& \numprint{37} &\numprint{37} & \numprint{38} &\numprint{38} &\numprint{38}& \numprint{38} \\
uk &8 &\numprint{73}& \numprint{73}& \numprint{73} &\numprint{73} & \numprint{77} &\numprint{77} &\numprint{76}& \numprint{76} \\
uk &16 &\numprint{133}& \numprint{133}& \numprint{134} &\numprint{134} & \numprint{136} &\numprint{136} &\numprint{137}& \numprint{138} \\
uk &32 &\numprint{223}& \numprint{224}& \numprint{225} &\numprint{226} & \numprint{233} &\numprint{236} &\numprint{240}& \numprint{241} \\
uk &64 &\numprint{360}& \numprint{362}& \numprint{364} &\numprint{366} & \numprint{387} &\numprint{388} &\numprint{402}& \numprint{403} \\
     \hline
whitaker3 &2 &\numprint{63}& \numprint{63}& \numprint{63} &\numprint{63} & \numprint{63} &\numprint{63} &\numprint{63}& \numprint{63} \\
whitaker3 &4 &\numprint{186}& \numprint{186}& \numprint{187} &\numprint{187} & \numprint{189} &\numprint{189} &\numprint{189}& \numprint{189} \\
whitaker3 &8 &\numprint{320}& \numprint{320}& \numprint{321} &\numprint{321} & \numprint{321} &\numprint{321} &\numprint{322}& \numprint{322} \\
whitaker3 &16 &\numprint{526}& \numprint{528}& \numprint{531} &\numprint{531} & \numprint{531} &\numprint{532} &\numprint{535}& \numprint{536} \\
whitaker3 &32 &\numprint{803}& \numprint{803}& \numprint{811} &\numprint{811} & \numprint{812} &\numprint{813} &\numprint{821}& \numprint{822} \\
whitaker3 &64 &\numprint{1183}& \numprint{1185}& \numprint{1198} &\numprint{1199} & \numprint{1213} &\numprint{1216} &\numprint{1233}& \numprint{1233} \\
     \hline
wing &2 &\numprint{626}& \numprint{626}& \numprint{628} &\numprint{628} & \numprint{700} &\numprint{702} &\numprint{704}& \numprint{704} \\
wing &4 &\numprint{1288}& \numprint{1290}& \numprint{1296} &\numprint{1297} & \numprint{1432} &\numprint{1433} &\numprint{1453}& \numprint{1456} \\
wing &8 &\numprint{1943}& \numprint{1949}& \numprint{1977} &\numprint{1980} & \numprint{2157} &\numprint{2170} &\numprint{2191}& \numprint{2191} \\
wing &16 &\numprint{2963}& \numprint{2995}& \numprint{3085} &\numprint{3095} & \numprint{3378} &\numprint{3388} &\numprint{3398}& \numprint{3405} \\
wing &32 &\numprint{4300}& \numprint{4315}& \numprint{4492} &\numprint{4513} & \numprint{4906} &\numprint{4917} &\numprint{4911}& \numprint{4921} \\
wing &64 &\numprint{5736}& \numprint{5778}& \numprint{6267} &\numprint{6287} & \numprint{6622} &\numprint{6644} &\numprint{6673}& \numprint{6677} \\
     \hline
wing\_nodal &2 &\numprint{410}& \numprint{410}& \numprint{410} &\numprint{410} & \numprint{415} &\numprint{415} &\numprint{415}& \numprint{415} \\
wing\_nodal &4 &\numprint{811}& \numprint{814}& \numprint{836} &\numprint{837} & \numprint{858} &\numprint{859} &\numprint{863}& \numprint{864} \\
wing\_nodal &8 &\numprint{1177}& \numprint{1184}& \numprint{1211} &\numprint{1216} & \numprint{1286} &\numprint{1287} &\numprint{1294}& \numprint{1294} \\
wing\_nodal &16 &\numprint{1793}& \numprint{1800}& \numprint{1841} &\numprint{1850} & \numprint{1965} &\numprint{1974} &\numprint{1985}& \numprint{1987} \\
wing\_nodal &32 &\numprint{2494}& \numprint{2500}& \numprint{2602} &\numprint{2625} & \numprint{2791} &\numprint{2800} &\numprint{2806}& \numprint{2807} \\
wing\_nodal &64 &\numprint{3243}& \numprint{3252}& \numprint{3460} &\numprint{3467} & \numprint{3683} &\numprint{3685} &\numprint{3696}& \numprint{3702} \\
        \hline
\end{tabular}
\caption{Detailed per instance results.}
\label{tab:detailedlongtimefour}
\end{table*}
\end{appendix}                                         

\end{document}